\documentclass[preprint,3p]{elsarticle}

\usepackage[utf8]{inputenc}
\usepackage{bm}
\usepackage{mathrsfs}
\usepackage{amsmath}
\usepackage{amsfonts}
\usepackage{amsthm}
\usepackage{multirow}
\usepackage{dsfont}
\usepackage{caption}
\usepackage{subfigure}
\usepackage{enumitem}
\usepackage{booktabs}
\usepackage{url}
\biboptions{sort&compress}

\usepackage{comment}

\usepackage[english]{babel}
\makeatletter
\newcommand{\mylabel}[2]{#2\def\@currentlabel{#2}\label{#1}}
\makeatother

\newtheorem{lemma}{Lemma}
\newtheorem{remark}{Remark}

\newcommand{\real}{\mathbb{R}}

\newcommand{\R}{\mathbb{R}}

\newcommand{\mcD}{\mathcal{D}}

\newcommand{\mcF}{\mathcal{F}}

\usepackage{xcolor}

\def\omg{{\Omega}}

\def \bb{\bm{b}}

\def \fb{\bm{f}}

\def \ab{\bm{a}}
\def \MMb{\bm{M}}

\def \Fb{\bm{F}}

\def \ub{\bm{u}}
\def \Ub{\bm{U}}

\def \vb{\bm{v}}
\def \xb{\bm{x}}

\def \zb{\bm{z}}

\def \yb{\bm{y}}

\def \eb{\bm{e}}

\def \Bb{\bm{B}}
\def \bb{\bm{b}}

\def \xib{{\boldsymbol\xi}}
\def \etab{{\boldsymbol\eta}}
\def \zetab{{\boldsymbol\zeta}}

\def \Xb{\bm{X}}

\def \Fb{\bm{F}}

\usepackage{amsmath}
\newcommand{\vertii}[1]{\lVert #1\rVert}

\newcommand{\Silling}[1]{{\color{orange}#1}}
\newcommand{\YY}[1]{{\color{blue}#1}}

\usepackage{lineno}

\usepackage{lipsum}
\usepackage{amsfonts}
\usepackage{graphicx}
\usepackage{epstopdf}
\usepackage{algorithm}
\usepackage{algorithmic}
\ifpdf
  \DeclareGraphicsExtensions{.eps,.pdf,.png,.jpg}
\else
  \DeclareGraphicsExtensions{.eps}
\fi

\usepackage{amsopn}

\journal{Computer Methods in Applied Mechanics and Engineering}
\begin{document}

\begin{frontmatter}

\title{Mesoscale Neural Operators: Learning Nonlinear Constitutive Laws from Molecular Dynamics for Scalable Fracture Predictions}

\author[lehigh]{Yue Yu\corref{cor1}}
\ead{yuy214@lehigh.edu}
\author[lehigh]{Jihong Wang}
\author[lehigh]{Stewart Silling}
\author[lehigh]{Ziwei Yang}

\cortext[cor1]{Corresponding author.}

\affiliation[lehigh]{organization={Department of Mathematics, Lehigh University},
             city={Bethlehem},
             postcode={18015},
             state={PA},
             country={USA}}


\begin{abstract}
Molecular dynamics (MD) simulations can capture the nanoscale physics of fracture at the atomistic level, but their computational cost prohibits predictions at the length scales and time scales needed for many applications. Machine learning approaches, on the other hand, have been proposed as heuristic surrogates for material fracture, yet they cannot guarantee the physical consistency of the predicted response. To address this shortcoming, we propose the mesoscale neural operator (MNO): a data-driven nonlocal constitutive model that is learned directly from MD loading and displacement data. After training, it can be used in continuum fracture simulations on much larger domains and grid sizes. The key idea of the MNO is to exploit the built-in characteristic length of a nonlocal model, called the horizon $\delta$, to travel across scales. Physical consistency across scales is obtained by building two structural ingredients into the MNO: a normalized kernel that renders the strain energy density invariant under horizon rescaling, and a rescaled critical bond stretch that keeps the energy release rate consistent across scales. Through tests on synthetic data, we show that the MNO and its learning workflow indeed discover the nonlocal continuum multiscale model in the form of the nonlocal kernel and the nonlinear constitutive relation. We then apply the framework to coarse-grained MD data and find that in a test problem, the learned model matches the MD strain energy density and critical energy release rate within $3\%$. It also predicts the peak transmitted forces within $6\%$ in out-of-distribution tension tests. As a further demonstration, the model is applied to a plate with $4000$ times the area of the training specimens and containing an initial defect. In this problem, the model reproduces the main features of dynamic brittle fracture, notably the transition from stable to unstable crack growth, including branch and cascading branch networks.
\end{abstract}



\begin{keyword}
Neural operators \sep Data-driven constitutive modeling \sep Fracture mechanics \sep  Peridynamics \sep Nonlocal models
\end{keyword}




\end{frontmatter}

\tableofcontents

\section{Introduction}

Fracture is an inherently multiscale phenomenon: cracks nucleate and grow when
atoms or molecules separate, yet the strength and failure modes
that matter in applications emerge at length scales and time scales thousands of times
larger.  Molecular dynamics (MD) simulations capture the microscale
physics of fracture directly, resolving each nanoscale event
without an externally imposed fracture criterion. However, because the
computational cost of MD confines simulations of fracture to very small volumes and time
durations, such models are difficult to use in many application-relevant
simulations.  Concurrent atomistic-to-continuum
coupling methods \cite{tadmor1996quasicontinuum,miller2002quasicontinuum,wagner2003coupling,xiao2004bridging,fish2007concurrent,parks2008connecting}
alleviate this cost by restricting the atomistic description to small
subregions, but they retain the atomistic resolution wherever cracks
may propagate and require delicate handshake regions between the two
descriptions.  An alternative strategy, which we pursue in this work,
is hierarchical: extract from MD data a coarse-grained
constitutive model that can then be scaled up and deployed in
continuum-style simulations on much larger domains
\cite{rudd1998coarse,seleson2009peridynamics}.

Machine learning approaches can be applied at the macroscale and
have been widely proposed as surrogates for material deformation and
fracture.  Model-free data-driven mechanics
\cite{kirchdoerfer2016data}
dispenses with a constitutive law altogether and has been extended to
fracture by representing the crack-growth resistance as a data set
\cite{carrara2020data}.
Sparse-regression approaches such as EUCLID discover interpretable
laws from full-field displacements and reaction forces alone,
including plasticity and generalized standard materials
\cite{flaschel2022discovering,flaschel2023automated}.
Variational system identification selects phase-field
fracture ingredients from synthetic full-field data
\cite{livingston2025inference}.
Neural constitutive and cohesive functions with no prescribed form
have been trained through differentiable solvers
\cite{huang2020learning}
or thermodynamically constrained losses \cite{wei2023data,masi2022multiscale}, and tensor-basis
architectures discover material anisotropy while enforcing frame invariance \cite{fuhg2022learning}
(see, e.g.,  \cite{fuhg2025review}
for a recent review).  Closer to our setting, coarse surrogates have
been learned from fine-scale simulations and deployed at larger
scales. These include neural cohesive laws extracted from MD and discrete-element
data \cite{fernandez2020application,wang2019meta}.
Recurrent surrogates with learned internal variables can replace
micromodels in multiscale (FE$^2$-type) simulations of softening,
plastic, and viscoelastic materials
\cite{ghavamian2019accelerating,liu2020deep,liu2022learning,liu2023learning,sohail2027neural}.
Crack-statistics emulators can inform continuum damage models
\cite{fernandez2021accelerating}. Stochastic multiscale models describe crack propagation in random
heterogeneous media \cite{hun2019stochastic}, and probabilistic learning has been used to enhance
fracture paths in random composites \cite{guilleminot2020data}.  Neural operators
\cite{lu2019deeponet,li2020fourier,li2020neural,karniadakis2021physics,lu2021comprehensive}
have also been applied to fracture, mostly in surrogate form by
learning the solution map of a given model
\cite{goswami2021physics,yin2022simulating,goswami2022deep,haghighat2020nonlocal,fan2022meshfree}.
However, these learned models remain largely heuristic surrogates
that cannot guarantee the physical consistency of the predicted
response.  

Fracture is dissipative and history dependent. The fields
localize or become discontinuous, and stress is almost never measured directly.
The damage variables used in typical continuum damage models are not observable, so the models are trained and
validated on displacements and boundary forces alone.  Nothing
ensures that the energetic quantities governing crack growth, the
stored energy and the energy release rate, are captured in learned models, especially
when the surrogate is deployed at a scale different from that of the
training data.  To the best of the authors' knowledge, no work in this line learns a fracture
constitutive law from MD displacements and forces and then runs it in
a continuum fracture simulation on a significantly larger domain.

To help overcome these current limitations in machine learning for fracture modeling, 
in this work we propose to learn a mesoscale constitutive
model in the form of a nonlocal neural operator.
The built-in characteristic
length in the model, which is called the {\emph{horizon}} $\delta$, 
provides the vehicle for traveling across scales. 
The design of the nonlocal neural operator uses bond-based peridynamics \cite{silling2000reformulation,silling2007peridynamic,silling_2010}, a continuum theory
that replaces the spatial derivatives of classical continuum mechanics by
an integral of pairwise bond forces over a neighborhood with radius $\delta$. 
As a result, the neural operator retains the main advantage of peridynamics for fracture modeling: displacement
discontinuities require no special treatment. Cracks nucleate,
propagate, and branch autonomously once a bond-breaking rule is
prescribed \cite{ha2011characteristics,agwai2011predicting,bobaru2015cracks,diehl2019review,javili2019peridynamics,madenci2018weak}.  

In the sense that it averages small-scale quantities over a larger length scale,
coarse-graining creates nonlocality in a model. The effective coarse-grained
interaction between material points obtained by averaging out
fine-scale degrees of freedom is nonlocal even when the fine-scale
physics is local \cite{silling2014origin,seleson2009peridynamics}.
In our prior work on machine learning of models from MD data, we
developed methods for learning the \emph{constitutive
operator} from data: nonlocal kernels identified from high-fidelity
data \cite{you2020data,you2022nonlocal,xu2022machine}.
Instead of choosing the nonlocal kernel and damage law
phenomenologically, 
peridynamic models are upscaled from coarse-grained MD of graphene
\cite{you2021md}.
This prior work has included a two-phase kernel-plus-damage-criterion
construction for MD data with fractures \cite{you2023towards}.
The peridynamic neural operator (PNO)  further guarantees
objectivity and momentum balance by construction
\cite{jafarzadeh2023peridynamic}. The PNO discovers material heterogeneity
from digital image correlation measurements
\cite{jafarzadeh2025heterogeneous,you2022learning,you2022physics}.
It admits conditionally unique solutions through a monotone
parameterization \cite{wang2026monotone}. The identifiability of such
kernels from displacement--load pairs is addressed in
\cite{yu2024nonlocal,wang2026rough}.

However, two challenges still remain before our earlier learned nonlocal models
can accomplish physically consistent fracture prediction across scales.  First,
the learned kernel and damage criterion are tied to the horizon and
resolution of the training data, whereas affordable large-scale
simulations require enlarging the horizon. Without additional
structure, such rescaling alters the stored energy and the energy
release rate, and hence the predicted strength and crack patterns.
Second, learned fracture models have so far been validated through
displacement errors, which do not necessarily gauge whether the energy balance
governing crack growth is also captured. While the data are provided only in the form of displacement--loading pairs, matching the energetic quantities becomes a learning problem without direct supervision. The energetic aspects are generally more challenging than reproducing the displacements. Yet they are necessary to guarantee robust downstream performance in loading scenarios and scales that lie outside the training distribution.

\begin{figure}[t!]
  \centering
  \includegraphics[width=\linewidth]{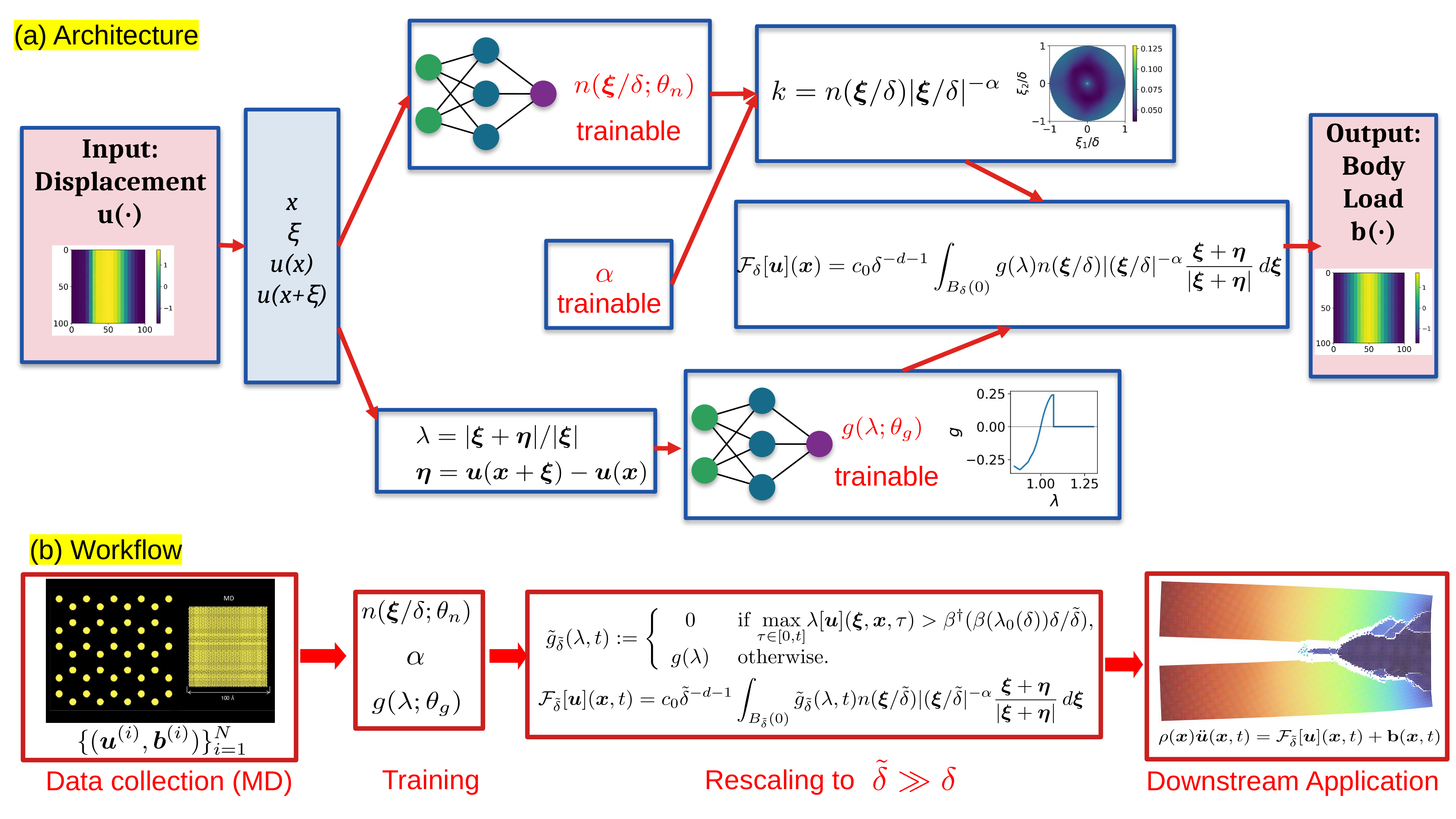}
  \caption{Overview of the proposed mesoscale neural operator (MNO) and the workflow of upscaling the characteristic length $\delta$.
  (a) Architecture: given a displacement field $\ub$, each bond $\xib$
  contributes to the nonlocal constitutive operator through two
  trainable networks and a trainable scalar: the kernel network
  $n(\xib/\delta;\theta_n)$, combined with the fractional exponent
  $\alpha$ into the singular kernel
  $k = n(\xib/\delta)\,|\xib/\delta|^{-\alpha}$, and the nonlinear
  constitutive network $g(\lambda;\theta_g)$ acting on the bond
  stretch $\lambda$.  Their product is integrated over the horizon
  $B_\delta(\bm{0})$ to return the body load $\bb$.  (b) Workflow: the
  constitutive operator is learned from coarse-grained MD
  displacement/body-load pairs $\{(\ub^{(i)},\bb^{(i)})\}_{i=1}^{N}$,
  rescaled from the training horizon $\delta$ to a larger horizon
  $\tilde\delta \gg \delta$ through the normalized kernel and the
  energy-consistent critical-stretch rule, and deployed in downstream
  dynamic fracture simulations at a much larger scale.}
  \label{fig:architecture}
\end{figure}

This work addresses both obstacles with the
\emph{mesoscale neural operator} (MNO), a normalized bond-based
constitutive operator whose strain energy density is invariant for affine deformation under
horizon rescaling. The method also rescales the critical stretch for bond
damage. This rescaling
keeps the energy release rate for a growing crack invariant across scales.  In
particular, our main contributions are as follows, with the main architecture and workflow provided in Figure \ref{fig:architecture}:
\begin{itemize}
\item We propose the mesoscale neural operator (MNO), a novel bond-based peridynamic model for material deformation and fracture prediction. By introducing the intrinsic length scale (the horizon size $\delta$) and normalizing the model with respect to it, the proposed MNO guarantees consistent energy density and energy release rate across different length scales for affine deformation.
\item A neural constitutive operator learning framework obtains the proposed mesoscale nonlocal operator directly from displacement/loading data pairs. Because a constitutive model rather than a solution map is learned, the resulting MNO is applicable to test samples with different resolutions, loading scenarios, and domain settings. The proposed normalization strategy also enables downstream simulations at a much larger scale.
\item We validate the learned nonlinear constitutive model through its energy density and energy release rate. In an exemplar application to the complex mechanical response of materials, our approach learns directly from quasi-static molecular dynamics simulations at the atomic length scale ($\approx10^{-10}$m). Comparing these two energy-based quantities of the learned model against those computed directly from the MD potential, the learned model attains physically consistent predictions (errors below $3\%$), while the predicted displacement fields on held-out samples remain accurate to within $10\%$.
\item We show that the nonlocal model captures the nonlinear responses and damage behaviors from data. To demonstrate this, in displacement-controlled tension tests outside the training distribution, the predicted peak transmitted forces match MD within $6\%$. We apply the model in downstream dynamic-fracture predictions at a much larger scale ($\approx 10^{-6}$m, with $4000$ times the area of the training specimens), and demonstrate its ability to reproduce the progression of brittle fracture morphologies. These include the transition under increasing load from the straight-ahead growth of a single crack to bifurcation and cascading branch networks.
\end{itemize}

The remainder of this paper is organized as follows.  In
Section~\ref{sec:MNO} we formulate the MNO, derive the normalization
that renders the strain energy density invariant across horizons and
the rescaled critical stretch that preserves the energy release rate,
and present the learning algorithm.
Section~\ref{sec:synthetic-verification} verifies the approach on
synthetic data sets generated at two characteristic lengths, where the
ground-truth constitutive law is known.  Section~\ref{sec:MD-application} applies the
framework to coarse-grained MD data. We describe the data generation,
learn a two-dimensional plane model, validate the learned model
through its energy density, energy release rate, and quasi-static
fracture tests against MD. We also deploy it in dynamic crack-branching
simulations with the horizon increased over the coarse-grained length scale
by a factor of 10.
Section~\ref{sec:conclusion} summarizes our findings and discusses
limitations and future directions.

\section{Mesoscale Neural Operator}\label{sec:MNO}

In this section, we formulate the proposed mesoscale neural operator (MNO) model, building on bond-based peridynamics \cite{wang2026monotone}.

\subsection{Bond-Based Peridynamics}

\textbf{Micropotential and pairwise force density.} Given a domain $\Omega\subset\R^d$, called the {\emph{reference configuration}}, the motion of a material point $\xb\in\Omega$ at time $t$ is governed by the following bond-based peridynamic equation:
\begin{equation}\label{eq:bbpd}
    \rho(\xb)\ddot\ub(\xb,t)= \int_{{B_\delta(\xb)}}\fb(\ub(\yb, t) - \ub(\xb, t), \yb-\xb)\;d\yb+\bb(\xb,t)\text{ ,}\quad (\xb,t)\in\omg\times[0,t_{\rm f}]\text{ ,}
\end{equation}
where $\ub$ is the displacement vector field, $\ddot\ub$ is the acceleration, $\bb$ is a prescribed body force density field, $\rho$ is the mass density in the reference configuration, and $\fb$ is the  {\emph{pairwise force density}}, whose value is the force vector (per unit volume squared) that the particle $\yb$ exerts on the particle $\xb$. The interaction between $\xb$ and $\yb$ is called a {\emph{bond}}. The relative position of the endpoints of the bond vector in the reference configuration is denoted by $\xib$ and their relative displacement by $\etab$:
\begin{equation}
\xib = \yb-\xb,~ \etab[\ub](\xib,\xb,t) = \ub(\xb+\xib, t)-\ub(\xb,t).
\end{equation}
The {\emph{bond stretch}} is defined as 
\begin{equation}
\lambda[\ub](\xib,\xb,t) = |\xib + \etab[\ub](\xib,\xb,t)|/ |\xib|, 
\end{equation}
which is the ratio of the deformed bond length to its undeformed length. It is assumed that $\lambda>0$ for all bonds, which means that two material points cannot overlap due to deformation. The energy per unit volume in the body at a given point (i.e., the local strain energy density) is given as
\begin{equation}\label{eq:energy_density}
    W(\xb,t) = \frac12\int_{{B_\delta(\bm{0})}}w(\lambda[\ub](\xib,\xb,t), \xib)\,d\xib,
\end{equation}
where $B_\delta(\bm{0})$ is the neighborhood of radius $\delta$ centered at $\bm{0}$, known as the {\emph{family}} of $\xb$. The positions in the volume integral are in the reference configuration. The stored energy in a bond under deformation, $w(\lambda, \boldsymbol{\xi})$, is called the {\emph{micropotential}}, which has dimensions of energy per unit volume squared and is non-negative. The pairwise force density is derived from $w$:
\begin{align*}
    \fb(\etab[\ub](\xib,\xb,t),\xib) = &\frac{\partial w}{\partial\etab}=\frac{\partial w}{\partial \lambda}(\lambda[\ub](\xib,\xb,t), \xib)\frac{\partial \lambda}{\partial \etab} \\
    =& \frac{\partial w}{\partial \lambda}(\lambda[\ub](\xib,\xb,t),\xib)\frac{1}{|\xib|} \MMb[\ub](\xib,\xb,t), \quad \forall \xib, \xb, t
\end{align*}
where the deformed bond direction vector is
\[
   \MMb[\ub](\xib,\xb,t) := \frac{\xib+\etab[\ub](\xib,\xb,t)}{|\xib+\etab[\ub](\xib,\xb,t)|}.
\]
Following \cite{wang2026monotone}, to enhance the identifiability and physical interpretability of microelastic models, we further assume that the contributions from bond extension and bond direction can be written in a separable form
\begin{equation}\label{eq:w}
 \frac{\partial w}{\partial \lambda}(\lambda,\xib)\frac{1}{|\xib|} =c_0 g(\lambda)k(\xib)=c_0 g(\lambda)n(\xib)|\xib|^{-\alpha}, \; \text{where }k(\xib)\geq 0,\, g(\lambda)\geq 0 \text{ for }\lambda\geq 1.
\end{equation}
Here, $c_0$ is a generic constant such that the physical units are consistent. Physically, Eq.~\eqref{eq:w} means that all the bond forces depend on the bond deformation in essentially the same way, but the forces could be lesser or greater for different bond lengths. It provides a clear physical interpretation of the learned model: $g$ captures the nonlinearity as the bond deforms, and $k$ characterizes the nonlocal interactions. When the material is undeformed, we have $g(1)=0$. Additionally, we require $g(\lambda)\geq 0$ for $\lambda\geq 1$ to impose the physical intuition that under extension the material generates an internal tensile force, which acts as a restoring force opposing the direction of deformation so that mechanical equilibrium can be attained. As will be discussed in Section \ref{sec:rescale_s0}, this constraint also guarantees the solvability of our proposed critical stretch rescaling law.

Denoting $\beta(\lambda):=\int_{1}^{\lambda} g(s) ds$ as the anti-derivative of $g$, the above equation yields a corresponding micropotential function:
$$w(\lambda[\ub](\xib,\xb,t),\xib)=c_0 n(\xib) |\xib|^{-\alpha+1} \beta(\lambda[\ub](\xib,\xb,t)).$$

\begin{remark}
In the MNO, the microstructural information is captured via a truncated fractional-order peridynamic kernel. As shown in \cite{benson2000application}, a visually homogeneous medium (a uniform sand pack) possesses sub-grid microstructural variations that induce non-Gaussian, heavy-tailed behavior. Instead of explicitly mapping these microscopic fluctuations, a spatial fractional-order operator natively encapsulates this sub-grid heterogeneity through power-law scaling. In a similar vein, a mesoscale solid domain exhibits sub-grid structural heterogeneities (e.g., micro-void distributions, grain boundary networks, or phase interfaces) that dictate non-classical stress distributions and non-local deformation fields. By combining a finite horizon ($\delta$) with a fractional-order kernel exponent ($\alpha$), our model establishes a rigorous, dual-mechanism representation of the microstructure. The horizon $\delta$ acts as a strict spatial truncation that defines the geometric bound of the mesoscopic interaction zone, preventing non-physical long-range effects. Meanwhile, the fractional order $\alpha$ governs the internal topology and rate of force attenuation within that zone, serving as a statistical proxy for the sub-grid heterogeneity.
\end{remark}

\textbf{Bond weakening and fracture.} In peridynamics, material damage is incorporated into the constitutive model by allowing the bonds between material points to break irreversibly. In particular, a scalar state function $\gamma[\ub](\xib,\xb,t)$ is defined and takes values in the interval $[0,1]$, to describe bond weakening and breakage as cracks grow:
\begin{equation}\label{eqn:gamma}
\gamma[\ub](\xib,\xb,t):=
\left\{\begin{array}{cl}
1 & \text{ if }\lambda[\ub](\xib,\xb,\tau)\leq \lambda_0,\, \forall \tau\in[0,t],\\
0 & \text{ otherwise,}\\
\end{array}\right.
\end{equation}
where $\lambda_0>1$ is the critical stretch criterion. $\lambda_0$ depends on the material.  $\gamma[\ub](\xib,\xb,t)$ is a history-dependent function, i.e., a bond can never recover once it exceeds the critical stretch criterion. When $\gamma[\ub](\xib,\xb,t)=1$, the bond between material points $\xb$ and $\xb+\xib$ is considered ``intact,'' and a change of the displacement at material point $\xb+\xib$ can directly affect the displacement at $\xb$. When the stretch $\lambda[\ub](\xib,\xb,\tau)$ exceeds the critical criterion $\lambda_0$ at some time $\tau<t$, the material is damaged and we have $\gamma[\ub](\xib,\xb,t)=0$ -- the bond between $\xb$ and $\xb+\xib$ is ``broken.''

\textbf{The classical bond-based peridynamic model.} In summary, the energy per unit volume for the bond-based peridynamic model defined by \eqref{eq:w} is given by
\begin{equation}\label{eq:energy_density_s}
    W(\xb,t) = \frac12 c_0\int_{{B_\delta(\bm{0})}} n(\xib) |\xib|^{-\alpha+1} \beta(\lambda[\ub](\xib,\xb,t))\gamma[\ub](\xib,\xb,t)\,d\xib,
\end{equation}
and the corresponding equation of motion reads
\begin{equation}\label{eq:model_0}
     \rho(\xb)\ddot\ub(\xb,t)=c_0\int_{B_{\delta}(\bm{0})} g(\lambda[\ub](\xib,\xb,t))n(\xib)|\xib|^{-\alpha} \gamma[\ub](\xib,\xb,t)\MMb[\ub](\xib,\xb,t) d\xib + \bb(\xb,t),
\end{equation}
for all $(\xb,t)\in\omg\times[0,t_{\rm f}]$, where $t_{\rm f}$ denotes the final simulation time. The model is closed with an initial condition and boundary data. The boundary data are prescribed in a boundary layer with finite volume, $\Omega_I=\{\xb|\xb\in\real^d \backslash \omg,\,\text{dist}(\xb,\omg) < \delta\}$, i.e.,
\begin{equation} \label{eq:bc}
\ub(\xb,t)=\ub_{BC}(\xb,t),\quad (\xb,t)\in\Omega_I\times[0,t_{\rm f}].
\end{equation}

\subsection{Normalized Kernel for Invariant Strain Energy Density}

In this and the next subsection, we consider a homogeneous body under isotropic extension:
$$\ub(\xb+\xib,t) - \ub(\xb,t)=s\xib,$$
where $s>-1$ is a constant for all $\xib$. Substituting it into the definition of $\lambda$, we notice that $\lambda = s+1$ in this class of deformations.

To obtain an invariant strain energy density formulation under change of $\delta$, we propose a normalized kernel formulation of $w$ as:
\begin{equation}\label{eqn:w}
w^{MNO}_\delta(\lambda,\xib)=c_0 \delta^{-d} n(\xib/\delta) |\xib/\delta|^{-\alpha+1} \beta(\lambda).
\end{equation}
Then the corresponding pairwise force function reads:
\begin{align*}
    \fb^{MNO}_\delta(\etab,\xib) = & \frac{\partial w^{MNO}_\delta}{\partial \etab}= \frac{\partial w^{MNO}_\delta}{\partial \lambda}(\lambda,\xib)\frac{1}{|\xib|} \frac{\xib+\etab}{|\xib+\etab|}\\
    =&c_0 \delta^{-d} n(\xib/\delta) |\xib/\delta|^{-\alpha+1} g(\lambda) \frac{1}{|\xib|} \frac{\xib+\etab}{|\xib+\etab|}\\
    =& c_0\delta^{-d-1} n(\xib/\delta) |\xib/\delta|^{-\alpha} g(\lambda) \frac{\xib+\etab}{|\xib+\etab|}
\end{align*}
for all $\xib \in B_\delta(\bm{0})$. We have the following property:
\begin{lemma}
For a homogeneous body under isotropic extension with all intact bonds, if the function $w^{MNO}_\delta(\lambda, \xib)$ is defined following \eqref{eqn:w}, then the corresponding energy functional $W^{MNO}_\delta$ is invariant with respect to the change of horizon size $\delta$.
\end{lemma}
\begin{proof}
Since all bonds are intact ($\gamma \equiv 1$), the energy functional reads:
\begin{align}
\nonumber    W^{MNO}_\delta(\xb,t) = &\frac12 \int_{{B_\delta(\bm{0})}}w^{MNO}_\delta(\lambda,\xib)\,d\xib\\
= &\frac12 \int_{{B_\delta(\bm{0})}}c_0 \delta^{-d} n(\xib/\delta) |\xib/\delta|^{-\alpha+1} \beta(\lambda[\ub](\xib,\xb,t)) \,d\xib.\label{eq:W}
\end{align}
Denoting $\zetab:=\xib/\delta$ and substituting $\lambda[\ub](\xib,\xb,t)=s+1$ into the above equation, we then have:
\begin{align}
\nonumber    W^{MNO}_\delta(\xb,t) = &\frac12 c_0 \beta(s+1)\int_{{B_1(\bm{0})}} n(\zetab) |\zetab|^{-\alpha+1} \,d\zetab,\label{eq:energy_density_rescale}
\end{align}
which is independent of $\delta$.
\end{proof}

\subsection{Rescaling the Critical Stretch for Consistent Energy Release Rate}\label{sec:rescale_s0}

Next, we rescale the damage criterion across different $\delta$ to ensure a consistent energy release rate. Even though the strain energy density formulation uses a normalized kernel, the damage criterion requires a modification of the nonlinear constitutive law characterized by the function $g$.

To derive this, we first describe how damage is implemented in the MNO. For a given horizon size $\delta$, we denote by $\lambda_0(\delta)$ the critical stretch for bond failure. The damage criterion is then incorporated into the MNO model by reformulating \eqref{eq:model_0} as:
\begin{equation}\label{eq:model_1}
     \rho(\xb)\ddot\ub(\xb,t)=c_0\delta^{-d-1}\int_{B_{\delta}(\bm{0})} \tilde{g}_\delta(\lambda[\ub](\xib,\xb,t),t)n(\xib/\delta)|\xib/\delta|^{-\alpha}\MMb[\ub](\xib,\xb,t) d\xib + \bb(\xb,t),
\end{equation}
where $\tilde{g}$ is defined as:
\begin{equation}
\tilde{g}_\delta(\lambda,t):=\left\{\begin{array}{cl}
0 & \text{ if }\max_{\tau\in[0,t]}\lambda[\ub](\xib,\xb,\tau) > \lambda_0(\delta),\\
g(\lambda) & \text{ otherwise}.
\end{array}\right.
\end{equation}


Now, we match the energy per unit fracture area for complete separation of the two halves of the body. Denote a spherical cap with height $z\leq {\delta}$ as $H_{\delta}(z)$. Since the energy {{per unit volume}} to break a single bond is equal to the micropotential function {{at the critical strain $\lambda_0$}}, the energy required to break all the bonds per unit fracture area is
\begin{align}\label{eq:G_def}
G_\delta(\lambda_0(\delta))=&\int_{0}^{{\delta}}\int_{H_{{\delta}}(z)}  w(\lambda_0(\delta),\xib)\, d\xib \, dz.
\end{align}
When upscaling the mesoscale model from $\delta$ to $\tilde{\delta}>\delta$, we propose the rescaling law for the critical stretch:
\begin{equation}\label{eqn:s}
\lambda_0(\tilde{\delta})=\beta^{\dag}(\beta(\lambda_0(\delta))\delta/\tilde{\delta}),
\end{equation}
where $\beta^{\dag}(l):= \min \{\lambda|\beta(\lambda)=l\}$. Since $g(\lambda)\geq 0$ for  $\lambda\geq 1$, we notice that its anti-derivative $\beta$ is monotone and $\beta(1)=0$. Hence, for any $l\in[0,\beta(\lambda_0(\delta))]$, there exists at least one solution of $\beta^{\dag}(l)$, and $\lambda_0(\tilde{\delta})$ is well-defined.

In the following, we show that the choice of \eqref{eqn:s} guarantees the physical consistency of fracture energy:
\begin{lemma}\label{lemma:2}
If the model is defined following \eqref{eq:model_1}, the micropotential function reads:
\begin{equation}\label{eqn:w_1}
w^{MNO}_\delta(\lambda,\xib,t)=c_0 \delta^{-d} n(\xib/\delta) |\xib/\delta|^{-\alpha+1} \tilde{\beta}_\delta(\lambda,t),
\end{equation}
with
\begin{equation}
\tilde{\beta}_\delta(\lambda,t):=\left\{\begin{array}{cl}
\int_1^{\lambda_0} g(s) ds  & \text{ if }\max_{\tau\in[0,t]}\lambda[\ub](\xib,\xb,\tau) > \lambda_0(\delta),\\
\int_1^\lambda g(s) ds & \text{ otherwise}.
\end{array}\right.
\end{equation}
Then, for a homogeneous body under isotropic extension, the rescaling law in \eqref{eqn:s} guarantees that the energy required to break all the bonds per unit fracture area, $G_\delta(\lambda_0(\delta))$, is invariant with respect to the change of horizon size $\delta$.
\end{lemma}
\begin{proof}
\eqref{eqn:w_1} is an immediate result from $\fb=\frac{\partial w}{\partial \etab}$ and the definition
\begin{equation}
 \fb^{MNO}_\delta(\etab,\xib,t)=c_0\delta^{-d-1}\tilde{g}_\delta(\lambda[\ub](\xib,\xb,t),t)n(\xib/\delta)|\xib/\delta|^{-\alpha}
\end{equation}
from \eqref{eq:model_1}. Substituting $\lambda=\lambda_0(\delta)$ and \eqref{eqn:w_1} into the definition in \eqref{eq:G_def}, we have
\begin{align}
\nonumber G^{MNO}_{{\delta}}(\lambda_0(\delta))=&c_0 {\delta}^{-d} \int_{0}^{{\delta}}\int_{H_{{\delta}}(z)}  n(\xib/{\delta}) |\xib/{\delta}|^{-\alpha+1} \beta(\lambda_0(\delta))\, d\xib \, dz \\
=&c_0 \beta(\lambda_0(\delta)) {\delta} \int_0^1 \int_{H_{1}(\varepsilon)}  n(\zetab) |\zetab|^{-\alpha+1} \, d \zetab \,d \varepsilon.\label{eq:G}
\end{align}
When upscaling the mesoscale model from $\delta$ to $\tilde{\delta}>\delta$, matching $G^{MNO}_{\tilde{\delta}}(\lambda_0(\tilde{\delta}))$ and $G^{MNO}_{{\delta}}(\lambda_0(\delta))$ at their critical stretch ratios then yields
$$G^{MNO}_{\tilde{\delta}}(\lambda_0(\tilde{\delta}))=G^{MNO}_{{\delta}}(\lambda_0(\delta)) \Rightarrow \beta(\lambda_0(\tilde{\delta})) \tilde{\delta}=\beta(\lambda_0(\delta)) {\delta},$$
which completes the proof.
\end{proof}


\subsection{Learning Algorithm}

As described above, our proposed mesoscale model is based on a peridynamic formulation with the following form:
$$\rho(\xb)\ddot{\ub}(\xb,t)=\mcF^{MNO}_{\delta}[\ub](\xb,t)+\bb(\xb,t),$$
where
\begin{equation}\label{eq:nonlocElasticity_comp}
\begin{aligned}
    \mcF^{MNO}_{\delta}[\ub](\xb,t):=&\int_{B_\delta (0)}   \fb^{MNO}_\delta(\etab,\xib,t)d\xib\\
    =&c_0\delta^{-d-1}\int_{B_\delta (0)}\tilde{g}_\delta(\lambda[\ub](\xib,\xb,t),t)n(\xib/\delta)|\xib/\delta|^{-\alpha}\MMb[\ub](\xib,\xb,t) d\xib \\
\lambda[\ub](\xib,\xb,t):=&\frac{|\xib+\etab[\ub](\xib,\xb,t)|}{|\xib|},\\
\tilde{g}_\delta(\lambda,t):=&\left\{\begin{array}{cl}
0 & \text{ if }\max_{\tau\in[0,t]}\lambda[\ub](\xib,\xb,\tau) > \lambda_0(\delta),\\
g(\lambda) & \text{ otherwise}.
\end{array}\right.
\end{aligned}
\end{equation}
When upscaling from $\delta$ to $\tilde{\delta}>\delta$, this framework guarantees the consistency of strain energy and energy release rate, by normalizing the kernel and the critical stretch ratio in this model:
$$\rho(\xb)\ddot{\ub}(\xb,t)=\mcF^{MNO}_{\tilde{\delta}}[\ub](\xb,t)+\bb(\xb,t),$$
where
\begin{equation}\label{eq:nonlocElasticity_comp_tilde}
\begin{aligned}
    \mcF^{MNO}_{\tilde{\delta}}[\ub](\xb,t):=&c_0\tilde{\delta}^{-d-1}\int_{B_{\tilde{\delta}} (0)}\tilde{g}_{\tilde{\delta}}(\lambda[\ub](\xib,\xb,t),t)n(\xib/\tilde{\delta})|\xib/\tilde{\delta}|^{-\alpha}\MMb[\ub](\xib,\xb,t) d\xib \\
\tilde{g}_{\tilde{\delta}}(\lambda,t):=&\left\{\begin{array}{cl}
0 & \text{ if }\max_{\tau\in[0,t]}\lambda[\ub](\xib,\xb,\tau) > \beta^{\dag}(\beta(\lambda_0(\delta))\delta/\tilde{\delta}),\\
g(\lambda) & \text{ otherwise}.
\end{array}\right.
\end{aligned}
\end{equation}

\textbf{Training procedure.} In practice, we construct the model with trainable horizon size $\delta$, kernel $n$, fractional order $\alpha$, and nonlinearity $g$ directly from displacement/body load data pairs. Without loss of generality, here we demonstrate the training on static/quasi-static datasets, and perform validation/test in both quasi-static and dynamic settings.

First, we parameterize the nonlinear function $g(\lambda)$ and kernel function $n(\xib/\delta)$ by two neural networks $g_{\theta_g}^{NN}$ and $n_{\theta_n}^{NN}$, respectively. To guarantee the well-posedness of the critical stretch ratio, we incorporate the non-negativity property of the function $g(\lambda)$ when $\lambda\geq 1$. Then, given discrete observations of function pairs:
\begin{eqnarray*}
\mathcal{D} = \{ (\ub^{(i)},\bb^{(i)})\}_{i=1}^{N} = \{ (\ub^{(i)}(\xb_j), \bb^{(i)}(\xb_j)): j=1,\cdots, N_x\}_{i=1}^{N}, 
\end{eqnarray*}
where $N$ is the number of samples, $\{ (\ub^{(i)},\bb^{(i)})\}$ represent pairs of real-valued continuous functions defined on a bounded, open, and connected domain  $\Omega\subset \R^d$, and $\chi = \{ \xb_j\in\Omega\}$ denotes the set of spatial mesh points. For each function pair and each bond $(\xb_k,\xb_j)$ with $\xb_k\in B_{\delta}(\xb_j)\cap \chi$, we then compute the corresponding bond stretch $\lambda^{(i)}_{j,k-j}:=|\xb_k-\xb_j+\ub^{(i)}(\xb_k)-\ub^{(i)}(\xb_j)|/|\xb_k-\xb_j|$, original bond $\xib_{j,k-j}:=\xb_k-\xb_j$, and deformed bond $\etab^{(i)}_{j,k-j}:=\ub^{(i)}(\xb_k)-\ub^{(i)}(\xb_j)$. The neural networks $g^{NN}$ and $n^{NN}$ take $\lambda^{(i)}_{j,k-j}$ and $\xib_{j,k-j}$ as their respective inputs. The integral in  \eqref{eq:nonlocElasticity_comp} is then approximated using a Riemann sum:
\begin{align}\label{eqn:Riemann_sum}
\mcF_{\delta,\Delta x}^{MNO,NN}[\ub^{(i)};\theta](\xb_j):=&\underset{\xb_k\in B_{\delta}(\xb_j)\cap \chi}{\sum} g_{\theta_g}^{NN}(\lambda^{(i)}_{j,k-j})n_{\theta_n}^{NN}(\xib_{j,k-j}/\delta) |\xib_{j,k-j}/\delta|^{-\alpha} \frac{\xib_{j,k-j}+\etab^{(i)}_{j,k-j}}{|\xib_{j,k-j}+\etab^{(i)}_{j,k-j}|}w_{jk}.
\end{align}
Here, $w_{jk}=\Delta x^d$ and $\Delta x$ is the  uniform grid size \footnote{Although we focus on the uniform grid here for simplicity, the model and learning procedure can also be trivially extended to cases with nonuniform grids by defining $w_{jk}$ as the optimized quadrature weights. See, e.g., \cite{fan2022meshfree,fan2021asymptotic}.}, and $d$ is the physical dimension of $\Omega$. 

To obtain the optimal model approximators $g^{NN}(\lambda)$, $n^{NN}(\xib/\delta)$ and $\alpha$, we split the dataset $\mcD$ into three non-overlapping subsets: 
training, validation, and test, respectively. Then we minimize the loss function with respect to the trainable parameter $\theta=\{{\theta_g},{\theta_n},
\alpha\}$ on the training dataset:
\begin{eqnarray}\label{eq:loss}
 \text{loss}(\theta) := \frac{1}{N_{train}}\sum_{i= 1}^{N_{train}}\dfrac{\vertii{\mcF_{\delta,\Delta x}^{MNO,NN}[\ub_{train}^{(i)};\theta]+\bb_{train}^{(i)}}_{l^2}}{\vertii{\bb_{train}^{(i)}}_{l^2}},
\end{eqnarray}
where $\vertii{\cdot}_{l^2}$ denotes the $l^2$ norm over the involved points. Notably, this learning algorithm exhibits resolution independence, as the prediction accuracy remains invariant with respect to the resolution of the input functions. In practice, we treat $\delta$ as a tunable hyperparameter, so as to mitigate the optimization challenges when optimizing $\delta$ simultaneously with other model parameters.

\textbf{Downstream rescaling and solution procedure.} Once the model is trained, the learned kernel and nonlinearity can be applied following \eqref{eq:nonlocElasticity_comp_tilde} for downstream simulations with a different domain shape, boundary setting, and length scale. Notably, $\lambda_0$ is calculated from the learned $g$ via:
\begin{equation}\label{eq:s0}
\lambda_0=\underset{\lambda\geq 1}{\text{argmax}} \,g(\lambda).
\end{equation}
We make this modeling choice based on the physical assumption that damage in bonds is irreversible. 
(The alternative assumption that bond damage can be repaired has also been used in other contexts \cite{kumar2024peridynamics}.)

To evaluate the error in $\ub$ in test samples, the learned model is solved given the test body force $\bb^{\text{test}}$ and the Dirichlet boundary condition $\ub(\xb,t) = \ub^{\text{test}}(\xb,t)$. On a given horizon scale $\tilde{\delta}$, we aim to solve for $\ub$ from the normalized governing equation:
$$\rho(\xb)\ddot{\ub}(\xb,t)=\mcF^{MNO}_{\tilde{\delta}}[\ub](\xb,t)+\bb^{\text{test}}(\xb,t),$$
and compare the numerical solution with the ground-truth solution $\ub_{\text{test}}(\xb,t)$. For temporal discretization, on $t^{n+1}:=(n+1)\Delta t$ a Newmark scheme is applied to update the approximation $\ub^{n+1}(\cdot)\approx \ub(\cdot,t^{n+1})$ (see, e.g., \cite{yu2012mixed}):
\begin{align}
\ub^{n+1} &= \ub^n+\Delta t\,\vb^n+\Delta t^2\!\left[(\tfrac12-\beta_{\rm nm})\ab^n+\beta_{\rm nm} \ab^{n+1}\right], \\
\vb^{n+1} &= \vb^n+\Delta t\!\left[(1-\gamma_{\rm nm})\ab^n+\gamma_{\rm nm}\ab^{n+1}\right], \\
\rho\,\ab^{n+1} &= \mcF_{\delta,\Delta x}^{MNO,NN}[\ub^{n+1};\theta] + \bb^{\text{test}}(\cdot,t^{n+1}).
\end{align}
Here we take 
$$\beta_{\rm nm}=1/4,\, \gamma_{\rm nm}=1/2$$
so that the scheme has second-order accuracy and unconditional stability in linear problems. $\mcF_{\delta,\Delta x}^{MNO,NN}[\ub;\theta]$ is Eq.~\eqref{eq:nonlocElasticity_comp} evaluated at $\ub$ with learned $\alpha$, $n$, and $g$.

Eliminating $\ab^{n+1}$ gives a nonlinear residual problem per step for
$\ub^{n+1}$:
\begin{align}
R(\ub^{n+1}) = &\frac{\rho}{\Delta t^2\beta_{\rm nm}}\left(\ub^{n+1} - \tilde{\ub}^{n+1}\right) - \mcF_{\delta,\Delta x}^{MNO,NN}[\ub^{n+1};\theta] - \bb^{\text{test}}(\cdot,t^{n+1}) = 0\label{eqn:Newton_residual}\\
\nonumber\tilde{\ub}^{n+1} = &\ub^n + \Delta t\,\vb^n + \Delta t^2(\tfrac12-\beta_{\rm nm})\ab^n.
\end{align}
This nonlinear system is solved using Newton's method, warm-started from the previous solution $\ub^n$.


We summarize the overall algorithm for training and evaluation of the proposed MNO model in Algorithm \ref{alg:1}. The algorithm comprises two primary components: (1) a training stage, where neural networks are trained to approximate the nonlinear constitutive relation \( g(\lambda) \) and kernel function \( k(\xib) \) while enforcing the physical constraints of Section~\ref{sec:rescale_s0} through the network architecture, and (2) a downstream application stage, where we employ the learned model to calculate the strain energy density and the critical energy release rate, and to solve for the displacement and crack evolution in a dynamic setting.  

\begin{algorithm} 
\caption{Learning and evaluation of MNO.} \label{alg:1}
\begin{algorithmic}[1]
\STATE \textbf{Inputs:} 
    \begin{itemize}
        \item Training, validation, and test datasets $\mcD_{train/valid/test}=\{(\ub^{(i)}_{train/valid/test}(\xb),\bb^{(i)}_{train/valid/test}(\xb))\}_{i=1}^{N_{train/valid/test}}$ on mesh points $\chi = \{\boldsymbol{x}_j\}_{j=1}^{N_x}$.
        \item Horizon size $\delta$ characterizing nonlocal interactions.
    \end{itemize}
\STATE \textbf{Pre-processing data:} 
\STATE 
Find and store the peridynamic neighbor node set: for each $\xb_j\in\chi$, find all $\xb_k\in\chi$ such that $|\xb_k - \xb_j|< \delta$.
\STATE \textbf{Training Stage:}
\FOR{$ep = 1: epoch_{max}$}
\FOR{each training batch}
    \STATE \begin{enumerate}
        \item Compute $\xib$, $\etab$, $\lambda$ and loss function \eqref{eq:loss}.
        \item Update parameters $\theta$ via Adam optimizer.
    \end{enumerate}
    \ENDFOR
\FOR{each validation batch}
        \STATE Compute $\xib$, $\etab$, $\lambda$ and the corresponding residual error.
    \ENDFOR
\STATE Store the optimal model parameter $\theta$ if the residual error reduces on the validation dataset $\mcD_{valid}$.
\ENDFOR
\STATE Evaluate $\lambda_0(\delta)$ following \eqref{eq:s0}, and set $g_\delta(\lambda)=0$ for $\lambda>\lambda_0(\delta)$.

\STATE \textbf{Downstream Validation and Application Stage:}
\STATE \begin{enumerate}
    \item \textbf{Validation: }For a given displacement/strain tensor profile, calculate the corresponding strain energy density functional $W_\delta^{MNO}$ following \eqref{eq:W}.
    \item \textbf{Validation: }With the calculated $\lambda_0(\delta)$, calculate the critical energy release rate $G_\delta^{MNO}$ following \eqref{eq:G}.
    \item \textbf{Application: }Rescale the model with respect to the target horizon size $\tilde{\delta}$ following \eqref{eq:nonlocElasticity_comp_tilde}.
    \item \textbf{Validation and Application: }For given domain ${\Omega}_{test}$ and loading $\boldsymbol{b}_{test}$, solve for the nonlinear model and obtain $\ub_{test}$.
    \end{enumerate}
\end{algorithmic}
\end{algorithm}

\section{Verification on Synthetic Data}
\label{sec:synthetic-verification}

Before turning to the coarse-grained molecular dynamics data, we use a
synthetic reference problem, in which the ground-truth constitutive law
is known, to verify whether the proposed MNO constitutive model can
represent the material response at more than one nonlocal length scale. In particular, we consider synthetic datasets corresponding to two characteristic lengths, $\delta$ and $2\delta$, which we denote as $S_\delta$ and $S_{2\delta}$, respectively:
\begin{equation}
\begin{aligned}
S_\delta:\, &(\Omega_\delta,h_\delta,\delta)
=([0,1]^2,1/128,0.125),\\
S_{2\delta}:\, &(\Omega_{2\delta},h_{2\delta},2\delta)
=([0,2]^2,1/64,0.25).
\end{aligned}
\label{eq:synthetic-configurations}
\end{equation}
Thus the domain side length, grid spacing, and physical horizon are all
doubled, while the relative nonlocal resolution is unchanged:
\begin{equation}
\frac{\delta}{h_\delta}
=\frac{2\delta}{h_{2\delta}}=16.
\label{eq:synthetic-resolution}
\end{equation}
For each length $\tilde{\delta}\in\{\delta,2\delta\}$, we consider an isotropic and static material model without fracture:
\begin{equation}\label{eqn:verification_model}
(\tilde{\delta})^{-d-1}\int_{B_{\tilde{\delta}} (0)}{g}(\lambda[\ub](\xib,\xb))n(|\xib/\tilde{\delta}|)|\xib/\tilde{\delta}|^{-\alpha}\MMb[\ub](\xib,\xb) d\xib+\bb(\xb)=0.
\end{equation}
Notice that $n(\xib/\tilde{\delta})=n(|\xib/\tilde{\delta}|)$ imposes the isotropic assumption. Moreover, since there is no fracture, we have $\lambda_0=\infty$ in \eqref{eq:nonlocElasticity_comp} and hence $\tilde{g}_\delta(\lambda)=g(\lambda)$. In our experiments we prescribe the ground-truth constitutive functions and parameter as
\begin{equation}
g_\star(\lambda)=\lambda-\lambda^{-3},
\qquad
n_\star(r)=\cos\!\left(\frac{\pi r}{2}\right),
\qquad
\alpha_\star=0.5,
\label{eq:synthetic-reference}
\end{equation}
and we denote $k_\star(r)=n_\star(r)r^{-\alpha_\star}$, $\psi_\star(\lambda,r)=g_\star(\lambda)k_\star(r)$ for notational simplicity in our verifications.

We then perform learning under two settings:
\begin{enumerate}
\item{\textit{Setting 1.}} Learning the constitutive model from the source dataset $S_\delta$ with characteristic length $\delta$, then evaluating the performance of the learned model on $S_{2\delta}$, a geometrically enlarged configuration corresponding to characteristic length $2\delta$. With this setting, we aim to verify the transferability of the MNO in downstream simulations at a larger scale.
\item{\textit{Setting 2.}} Learning the constitutive model from $S_\delta \bigcup S_{2\delta}$ with observations at
both scales, then evaluating the performance of the learned model on test samples at both scales. With this setting, we aim to investigate the capability of the MNO in handling data from multiple scales.
\end{enumerate}
For both settings, we evaluate the prediction error on a set of held-out samples, as well as the constitutive operator errors obtained by comparing the learned $g$, $n$, $\alpha$, $k$, and $\psi$ with the corresponding ground truths.

\subsection{Synthetic Data Generation and Training}
\label{subsec:synthetic-data}

To generate the training, validation, and test samples in $S_\delta$ and $S_{2\delta}$, we follow a practice similar to that in \cite{wang2026rough,wang2026monotone}: for each $\tilde{\delta}\in\{\delta,2\delta\}$, we first randomly sample multiple realizations of displacement fields $\ub$, then generate the corresponding force field by substituting $\ub$ into \eqref{eqn:verification_model} and evaluate $\bb$ numerically on a refined mesh. In particular, for the $i$-th realization, displacement component
$\ell\in\{1,2\}$, and nominal maximum absolute relative bond-stretch level
$s\in\mathcal S=\{0.02,0.05,0.10,0.15\}$, the displacement vector field $\ub^{(i)}(x,y)$ is generated  with the following procedure: we first sample $a_{\ell, j}^{(i)}\sim\mathcal{U}[-1,1]$ and $b_{\ell, j}^{(i)}\sim\mathcal{U}[-1,1]$ for $j=0, \cdots, 39$, and assemble them into a non-normalized vector field:
\begin{equation}
\bm{u}_0^{(i)}=\left(\begin{array}{c}
\sum_{j=0}^{39}e^{-j/40}
\left[
a_{1, j}^{(i)}\sin(j\pi x)
+b_{1, j}^{(i)}\cos(j\pi y)
\right]\\
\sum_{j=0}^{39}e^{-j/40}
\left[
a_{2, j}^{(i)}\sin(j\pi x)
+b_{2, j}^{(i)}\cos(j\pi y)
\right]
\end{array}\right).
\end{equation}
Then, we generate the normalized displacement field as:
\begin{equation}
\ub^{(i)}(x,y)
=A_{i,s}\bm{u}_0^{(i)}(x,y),
\label{eq:synthetic-fourier-data}
\end{equation}
where the positive
multiplier $A_{i,s}$ is defined via
\begin{equation}
\max_{\xb_j\in \chi,\,\xb_k\in B_{\tilde{\delta}}(\xb_j)\cap \chi}
\left|\lambda[A_{i,s}\bm u_0^{(i)}]
(\xb_k-\xb_j,\xb_j)-1\right|=s.
\label{eq:synthetic-amplitude-calibration}
\end{equation}
This guarantees that the maximum bond-stretch level reaches the pre-defined level $s$. Once $\ub^{(i)}$ is obtained, the corresponding body load is generated via:
$$\bb_{\tilde{\delta}}^{(i)}(\xb)=-(\tilde{\delta})^{-d-1}\int_{B_{\tilde{\delta}} (0)}{g_\star}(\lambda[\ub^{(i)}](\xib,\xb))n_\star(|\xib/\tilde{\delta}|)|\xib/\tilde{\delta}|^{-\alpha_\star}\MMb[\ub^{(i)}](\xib,\xb) d\xib.$$
For each $\tilde{\delta}$, we generate training, validation, and test sets containing $100$, $20$, and $80$ sample function pairs, respectively, and denote the corresponding datasets as $S_{\tilde{\delta},train}$, $S_{\tilde{\delta},valid}$, and $S_{\tilde{\delta},test}$.  


In this example, we enforce the following constraints on $g$ and $n$:
$$g(1)=0,\, g(\lambda)\geq 0,\,n(0)=1.$$
The first two conditions, as discussed in Section \ref{sec:rescale_s0}, aim to ensure the well-posedness of the critical stretch rescaling law. The last constraint is a normalization convention that does not restrict the constitutive response, since any positive amplitude factor in $n$ can be absorbed into $g$. In practice, we parameterize the constitutive functions as
\begin{equation}
\begin{aligned}
g_\theta(\lambda)
&=(\lambda-1)\operatorname{softplus}\!\left(N_g(\lambda-1)\right),\\
n_\theta(r)&=1+r^2N_n(r^2),\\
\alpha_\theta&=a,
\end{aligned}
\label{eq:synthetic-network-parameterization}
\end{equation}
where $a$ is an unconstrained trainable scalar initialized at $0.2$.
$N_g$ and $N_n$ are multi-layer perceptrons (MLPs) with four ReLU hidden layers of width 64 and three $\tanh$ hidden layers of width 16, respectively.

In Setting 1, we perform training on the samples in $S_{\delta,train}$, while in Setting 2 the training is on $S_{\delta,train}\bigcup S_{2\delta,train}$. To define the training, validation, and test losses, we denote: 
\begin{equation}\label{eq:synthetic-force-metrics}
\begin{split}
&E_{\tilde{\delta},train/valid/test}
:=\left[
\frac{1}{\#(S_{\tilde{\delta},train/valid/test})}\sum_{(\ub,\bb)\in S_{\tilde{\delta},train/valid/test}}\dfrac{\vertii{\mcF_{\tilde{\delta},\Delta x}^{MNO,NN}[\ub;\theta]+\bb}_{l^2}}{\vertii{\bb}_{l^2}}
\right]^{1/2},\\
&E_{\mathrm{comb},train/valid/test}
:=\left(\frac{(E_{\delta,train/valid/test})^2+(E_{2\delta,train/valid/test})^2}{2}\right)^{1/2}.
\end{split}
\end{equation}
where $\#(S)$ is the number of samples in set $S$. In Settings 1 and 2, we perform training on $E_{\delta,train}$ and $E_{\mathrm{comb},train}$, respectively:
\begin{eqnarray}\label{eq:loss_1}
 \theta_1:=\{{\theta_{1,g}},{\theta_{1,n}},
\alpha_1\}=\text{argmin} \; E_{\delta,train},\\
\theta_2:=\{{\theta_{2,g}},{\theta_{2,n}},
\alpha_2\}=\text{argmin} \; E_{\mathrm{comb},train},
\end{eqnarray}
and select the optimal parameter sets via the corresponding validation losses. Both models are trained with AdamW for 800 epochs under the same training schedule.

\subsection{Error Evaluations and Results}
\label{subsec:synthetic-evaluation}

To evaluate the training results in both settings, we consider two metrics. First, in Table~\ref{tab:synthetic-results} part (a) we report the prediction errors on the held-out test samples in the form of $E_{\tilde{\delta},train}$, $E_{\tilde{\delta},test}$, and $E_{\mathrm{comb},train}$. This part aims to provide a direct evaluation of the model performance in new and unseen loading scenarios with different characteristic lengths. Second, in Table~\ref{tab:synthetic-results} part (b) we demonstrate the operator errors by comparing the learned
constitutive functions against the ground-truth functions. Denoting the learned functions from our two settings with parameters $\theta_\ell$, $\ell=1,2$, as 
$$g_{\ell}(\lambda),\; n_{\ell}(r), \; k_{\ell}(r)=n_{\ell}(r)|r|^{-\alpha_{\ell}},\; \psi_{\ell}(\lambda,r)=g_{\ell}(\lambda)n_{\ell}(r)|r|^{-\alpha_{\ell}},$$
we evaluate the relative $L^2$ errors on their definition domains\footnote{The
lower endpoint $1/16$ of $D_k$ is the smallest nonzero normalized bond radius
on either stencil and excludes the singular point of
$k_\star(r)=n_\star(r)r^{-\alpha_\star}$.}
$D_g=[0.85,1.15]$, $D_n=[0,1]$, $D_k=[1/16,1]$, and $D_\psi=D_g\times D_k$ as
\begin{equation}
\label{eq:synthetic-constitutive-metric}
\begin{split}
&E_\alpha = \frac{|\alpha_{\ell}-\alpha_{\star}|}{\alpha_{\star}},\, E_g
=\frac{\|g_{\ell}-g_\star\|_{L^2(D_g)}}
{\|g_\star\|_{L^2(D_g)}},\, E_n
=\frac{\|n_{\ell}-n_\star\|_{L^2(D_n)}}
{\|n_\star\|_{L^2(D_n)}},\\
&E_k
=\frac{\|k_{\ell}-k_\star\|_{L^2(D_k)}}
{\|k_\star\|_{L^2(D_k)}},\, 
E_\psi
=\frac{\|\psi_{\ell}-\psi_\star\|_{L^2(D_\psi)}}
{\|\psi_\star\|_{L^2(D_\psi)}}.
\end{split}
\end{equation}
Intuitively, $E_g$, $E_n$, $E_k$, and $E_\psi$ measure the errors of bond law, bounded radial
profile, complete radial kernel, and complete bond force, respectively. This part provides direct evaluations on the capability of our proposed learning algorithm in discovering the true material constitutive law.


\begin{table}[H]
\centering
\caption{Verification on synthetic data: prediction and operator errors}
\label{tab:synthetic-results}
\setlength{\tabcolsep}{3pt}
\begin{tabular}{@{}l|lcrrr@{}}
\toprule
\multicolumn{6}{l}{(a) Prediction Errors} \\
\midrule
\addlinespace[3pt]
& Data set & Number of Samples
& $E_\delta$ & $E_{2\delta}$ & $E_{\mathrm{comb}}$ \\
\midrule
\multirow{3}{*}{Setting 1: Training on $S_\delta$}
& Training   & 100 & $0.0104\%$ & $0.1040\%$ & $0.0738\%$ \\
& Validation & 20  & $0.0101\%$ & $0.1030\%$ & $0.0729\%$ \\
& Test       & 80  & $0.0104\%$ & $0.0995\%$ & $0.0707\%$ \\
\midrule
\multirow{3}{*}{Setting 2: Training on $S_\delta$ and $S_{2\delta}$}
& Training   & 100 & $0.0086\%$ & $0.0124\%$ & $0.0106\%$ \\
& Validation & 20  & $0.0082\%$ & $0.0117\%$ & $0.0101\%$ \\
& Test       & 80  & $0.0084\%$ & $0.0122\%$ & $0.0105\%$ \\
\bottomrule
\end{tabular}
\medskip
\begin{tabular}{@{}l|ccccc@{}}
\toprule
\multicolumn{6}{l}{(b) Operator Errors} \\
\midrule
\addlinespace[3pt]
& $E_\alpha$ & $E_g$ & $E_n$ & $E_k$ & $E_\psi$ \\
\midrule
Setting 1: Training on $S_\delta$
& 3.95\% & 4.19\% & 2.62\% & 4.38\% & 0.58\% \\
Setting 2: Training on $S_\delta$ and $S_{2\delta}$
& 1.19\% & 1.59\% & 1.01\% & 1.49\% & 0.41\% \\
\bottomrule
\end{tabular}
\end{table}

\begin{figure}[H]
\centering
\includegraphics[width=\linewidth]{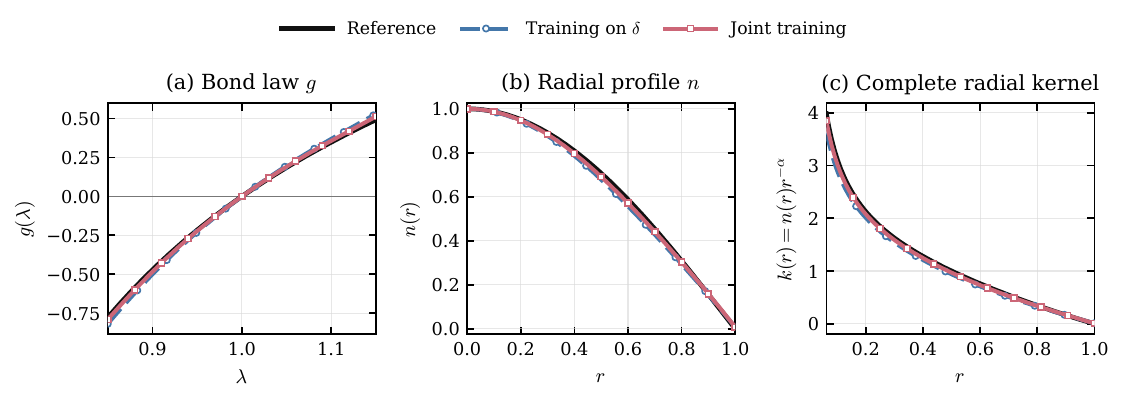}
\caption{Verification on synthetic data: comparison between the ground-truth functions and learned functions from Setting 1 (trained on a single length scale $S_\delta$) and Setting 2 (trained on multiple length scales $S_\delta$ and $S_{2\delta}$) in (a) bond law $g(\lambda)$;
(b) normalized radial profile $n(r)$; and (c) complete radial kernel
$k(r)=n(r)r^{-\alpha}$.}
\label{fig:synthetic-constitutive-functions}
\end{figure}

From Setting 1, one can see that a constitutive relation learned on $S_{\delta}$ can be applied without recalibration to the $2\delta$ samples with a mild increase of test errors from
$E_{\delta,test}=0.0104\%$ to $E_{2\delta,test}=0.0995\%$, providing numerical evidence that the proposed normalization carries the learned material relation to a geometrically enlarged configuration. In Setting 2, the training is performed on both characteristic lengths. The error on the original configuration $S_\delta$ is slightly decreased from $0.0104\%$ to $0.0084\%$, while reducing the error on the enlarged configuration $S_{2\delta}$ from $0.0995\%$ to $0.0122\%$. This result suggests that including data from multiple scales in training helps to improve the prediction accuracy after scale-up without reducing the accuracy at the original scale. In part (b), the operator errors show the same overall trend: while training on $S_\delta$ alone discovers the ground-truth bond law, radial profile, complete radial kernel, and fractional exponent with $\leq 5\%$ errors, learning from both characteristic lengths further reduces the errors to $\leq 2\%$. As shown in Figure~\ref{fig:synthetic-constitutive-functions}, the corresponding learned constitutive functions are highly consistent with the ground truths in both settings. Overall, these tests verify that:
\begin{enumerate}
\item The proposed mesoscale constitutive relation learned from a small characteristic length ($\delta$) can be applied to an enlarged length scale ($2\delta$) with a small ($\approx 0.1\%$) prediction error. 
\item The consistency across lengths also enables learning from multiple length scales, which further improves the accuracy of both the prediction and the constitutive operator across the scale-up test.
\end{enumerate}







\section{Application: Data-Driven Model from Molecular Dynamics}\label{sec:MD-application}

\begin{figure}
\centering
\raisebox{-0.5\height}{\includegraphics[width=0.5\textwidth]{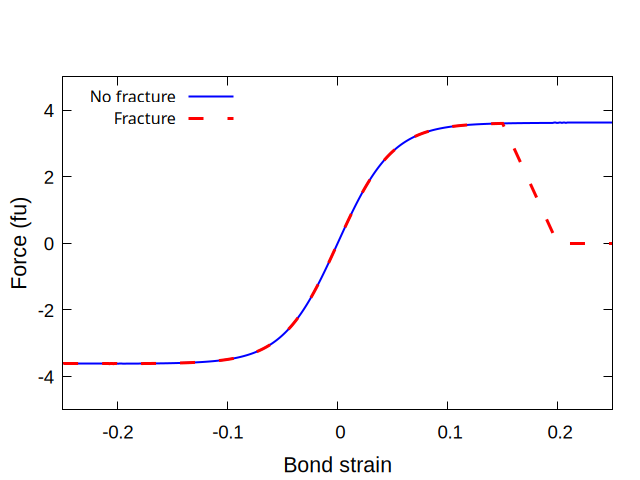}}\hfill
\raisebox{-0.5\height}{\includegraphics[width=0.47\textwidth]{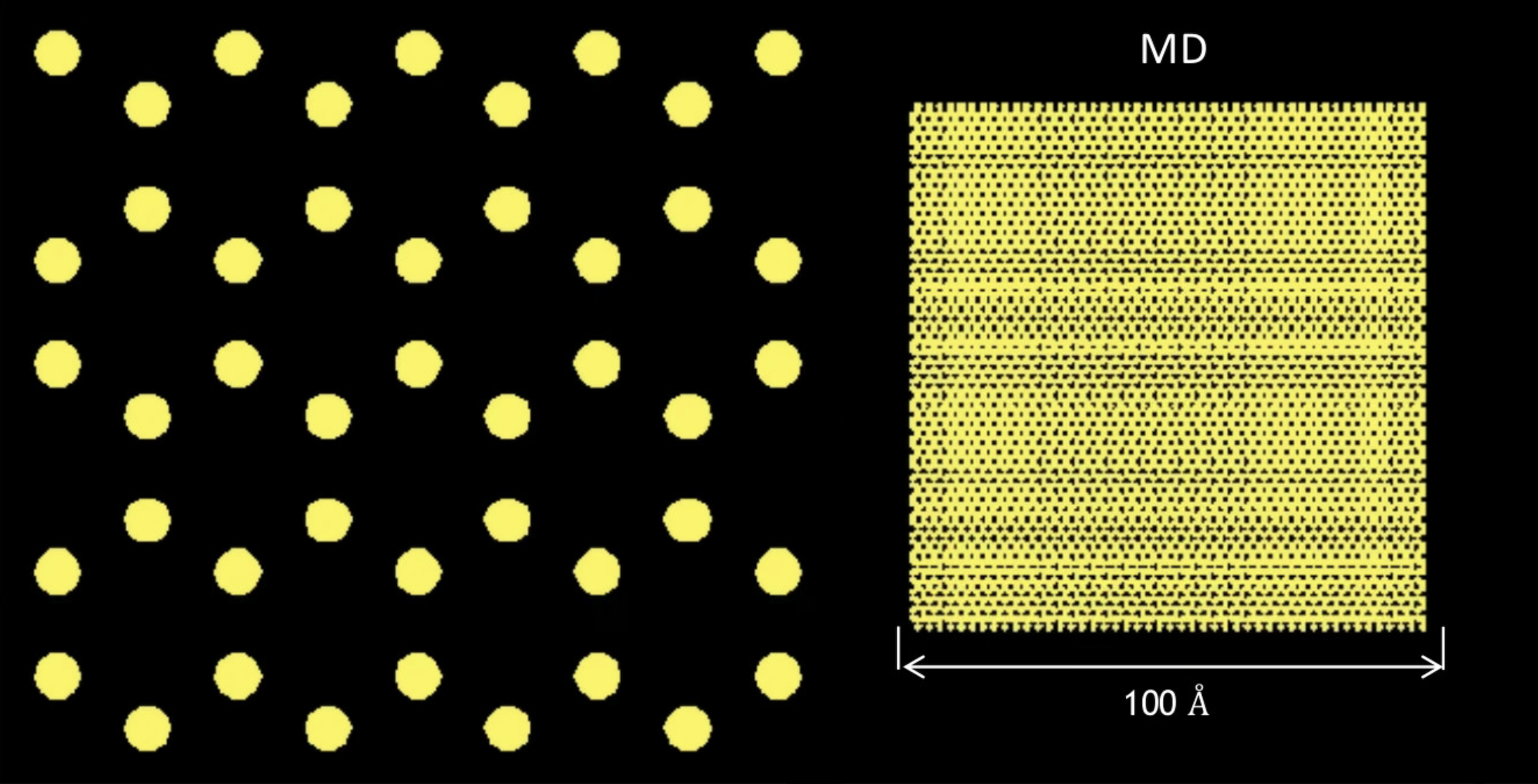}}
\caption{Settings in the MD simulations. Left: force between two particles vs. bond strain. Right: hexagonal atomic structure and simulation domain.
}
\label{fig-2pt}
\end{figure}

To illustrate the efficacy of our method, in this section we apply the learning approach to obtain an optimal peridynamic model from coarse-grained molecular dynamics (MD) displacements. The MD dataset, the pretrained model, and all codes are available in our public repository \cite{PNOgithub}.

\subsection{Data Generation}\label{sec:MD}

The machine learning algorithm is trained using MD data, consisting of displacements and body forces from $2$D MD simulations. To generate the training and validation datasets, we consider a nonlinear interparticle pair potential. The following system of units is used as a convenience in the MD model parameters and output data: 
\begin{itemize}
\item Length units ($lu$): $1\,lu=1\times 10^{-10}\,m$
\item Mass units ($mu$): $1\,mu=1\times 10^{-27}\,kg$
\item Time units ($tu$): $1\,tu=1\times 10^{-12}\,s$
\item Force units ($fu$): $1\,fu=1\times 10^{-13}\,N = 1\times 10^{-13}\,kg\cdot m/s^2$
\item Energy units ($eu$): $1\,eu=1\times 10^{-23}\,J = 1\times 10^{-23}\,kg\cdot m^2/s^2$
\item Stress units ($su$): $1\,su=1\times 10^{7}\,Pa = 1\times 10^{7}\,kg/m/s^2$
\end{itemize}

The particles are initially in a hexagonal lattice with initial position vectors
$\Xb_1,\Xb_2,\dots$.
The displacement vectors, which are time-dependent, are denoted by
$\Ub_1,\Ub_2,\dots$.
The initial lattice spacing is $1.40\,lu$.
Two particles interact if their bond length (initial separation distance) is less than a
cutoff distance $D$, where $D=4.242\,lu$.
The particle mass is $19.9\,mu$. See Figure \ref{fig-2pt} for the atomic structure and computational domain in MD settings.

We then apply force loadings on MD atoms and evolve the resulting displacement fields. Let $i$ and $j$ be two distinct particles.
Define the initial interatomic bond length $r$, extension $e$, and strain $p$ by
\[  
   r=|\Xb_j-\Xb_i|,\; e=|(\Xb_j+\Ub_j)-(\Xb_i+\Ub_i)|-r,\; p=\frac{e}{r}.
\]
When $r\le D$, the force between two particles is modeled as a nonmonotonic (nonconvex) function that can form fractures under some loading conditions:
\[
  F=\frac{A}{p_0}\tanh\left(\frac{p}{p_0}\right)\mu(p)
\]
where
\[
  \mu(p)=\left\{   \begin{array}{ll}
  1                 & {\mathrm{if}}\;p<p_*, \\
  1-(p-p_*)/p_0       & {\mathrm{if}}\;p_*\le p\le p_*+p_0, \\
  0                 & {\mathrm{if}}\;p_*+p_0<p   \end{array}
  \right.
  \]
and $p_*=0.15$. If the bond strain exceeds $p_*$, the force decreases as the strain is increased further.
This only happens in tension, not compression, as shown in the left plot of Figure~\ref{fig-2pt}. Additionally, we assume that the force vector $\Fb_{ji}$ is always directed parallel to the deformed bond vector, thus
the force that $j$ exerts on $i$ is
\[
  \Fb_{ji}=\frac{(\Xb_j+\Ub_j)-(\Xb_i+\Ub_i)}{r+e}\,F,
\]
from which it is immediate that $\Fb_{ij}=-\Fb_{ji}$.

The simulations model a hexagonal lattice of particles $i$ loaded by
spatially periodic external forces of the form
\[
  \Bb_i = Ct\,\sin(\pi \boldsymbol{\omega}\cdot\Xb_i/20)\,\eb
\]
where $C$ is a constant, $\boldsymbol{\omega}$ is a vector, and $\eb$ is a unit vector
that takes on the values either $(1,0)$ or $(0,1)$.
$C$ is small enough that the MD simulations are essentially quasi-static.
The training data consists of separate files for each combination of $\boldsymbol{\omega}$ and $\eb$.
A file contains the MD simulation data $\Ub_i(t^n),\Bb_i(t^n)$ for all the particles $i$
for a sequence of times $t^n$ as the loading is increased over time. $24$ such trajectories were generated, covering $\boldsymbol{\omega}$ magnitudes in $\{1.0, 1.414, 2.0, 2.236, 2.828, 3.0, 3.162\}$, for the purpose of training, validation, and test. Before being used for training, the MD data in each of the files is coarse-grained onto a coarser spatial discretization, which is a $21\times21$ grid. At each time $t^n$, at each node in the coarser discretization, the coarse graining (CG) algorithm aggregates the nearby MD displacements and external loads. The aggregation at each CG node computes the mass-weighted mean MD values within a radius $R$, yielding CG displacements and external loads $\ub$ and $\bb$ on the CG grid.

\begin{figure}  
\centering
\includegraphics[width=0.7\textwidth]{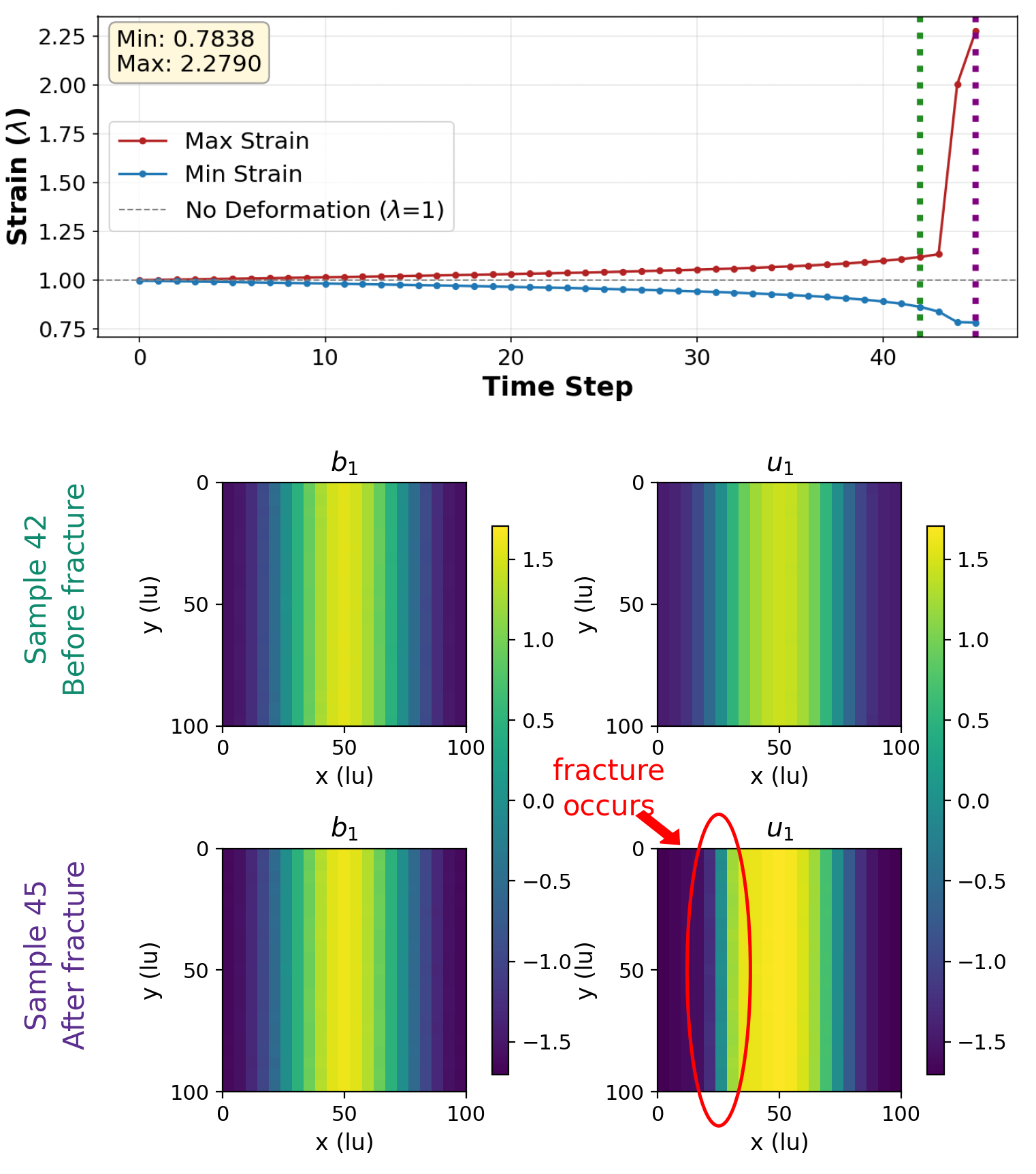}
\caption{Exemplar coarse-grained MD simulation result corresponding to one sequence, with linearly increasing force loading magnitude and correspondingly increasing bond stretch ratio ($\lambda$). The force loading and corresponding displacement fields are plotted at two time instances: one before fracture occurs (Sample 42) and one with fracture (Sample 45). In the latter, a vertical fracture is visible where the blue and yellow particles meet.}
\label{fig-atoms}
\end{figure}

In Figure \ref{fig-atoms} we demonstrate the trajectory corresponding to $\boldsymbol{\omega}=[1,0]$. There are $46$ time instances (samples) in this file, with zero loading at the beginning and linearly increasing loading with increasing sample index. In the upper plot of Figure \ref{fig-atoms} we show the minimum and maximum bond stretch ratios ($\lambda$) as functions of the sample index. One can see that the sample deforms more as the loading increases. Starting from the $43$rd sample, the maximum value of $\lambda$ exceeds $1.15$, and a slight increase of the body load induces substantial deformation. This is because fracture starts to form once the MD bond strain exceeds $0.15$, as illustrated by the force-vs-bond-strain curve in the left plot of Figure \ref{fig-2pt}. In the bottom plot of Figure \ref{fig-atoms} we illustrate the body load and corresponding displacement fields of sample indices $42$ and $45$. There is a visible jump of displacement in sample $45$, indicating a vertical fracture in the object. We perform a similar procedure on all sampled files corresponding to different $\boldsymbol{\omega}$, creating $800$ coarse-grained $(\bb(\xb),\ub(\xb))$ function pairs. All resulting function pairs are treated as quasi-static samples, and randomly split into three sets: $200$ samples in a training dataset, $50$ samples in a validation dataset, and $550$ samples in a held-out test dataset.





\subsection{Learning A 2D Plane Model}\label{section:MDlearn}
Since the MD simulations are restricted to 2D, the same assumption is applied
in the MNO. Because the peridynamic model uses volume
integrals to represent the forces between material points, the system is
assigned an arbitrary thickness $H$, with $H\ll\delta$. To perform learning on
the described MD data, we therefore reformulate the bulk MNO model of Section
\ref{sec:MNO} in a consistent 2D plane setting. Since $H\ll \delta$, the bond
force $\fb_\delta^{MNO}$ in \eqref{eq:nonlocElasticity_comp} can be considered
invariant along the thickness direction (denoted as $h$), and our goal is to
obtain a constitutive model which remains physically consistent across
different characteristic lengths $\delta$ in the other two directions (denoted
as $\xb=(x,y)$). As a result, we define all quantities on a $d=2$ domain, and
reformulate the MNO model as:
$$\rho(\xb)\ddot{\ub}(\xb,t)=\mcF^{MNO,2D}_{\delta}[\ub](\xb,t)+\bb(\xb,t),$$
where
\begin{equation}\label{eq:nonlocElasticity_comp_2d}
\begin{aligned}
    \mcF^{MNO,2D}_{\delta}[\ub](\xb,t):=&\int_0^H\int_{B^{2D}_\delta (0)}  \fb^{MNO,2D}_\delta(\etab,\xib,t) d\xib dh\\
  =&c_0\delta^{-d-1}H \int_{B^{2D}_\delta (0)}\tilde{g}_\delta(\lambda[\ub](\xib,\xb,t),t)n(\xib/\delta)|\xib/\delta|^{-\alpha}\frac{\xib+\etab[\ub](\xib,\xb,t)}{|\xib+\etab[\ub](\xib,\xb,t)|} d\xib.
  \end{aligned}
\end{equation}
Herein, $d\xib$ has dimensions of volume in 3D but area in 2D, that is, length$^d$. As a result, the units of $\rho$, $\ub$, and $\bb$ are $mu/lu^3$, $lu$, and $fu/lu^3=mu/lu^2/tu^2$, respectively. The units of $c_0$ are $fu/lu^3=mu/lu^2/tu^2$.

\begin{figure}[H]
\centering
\includegraphics[width=0.55\linewidth]{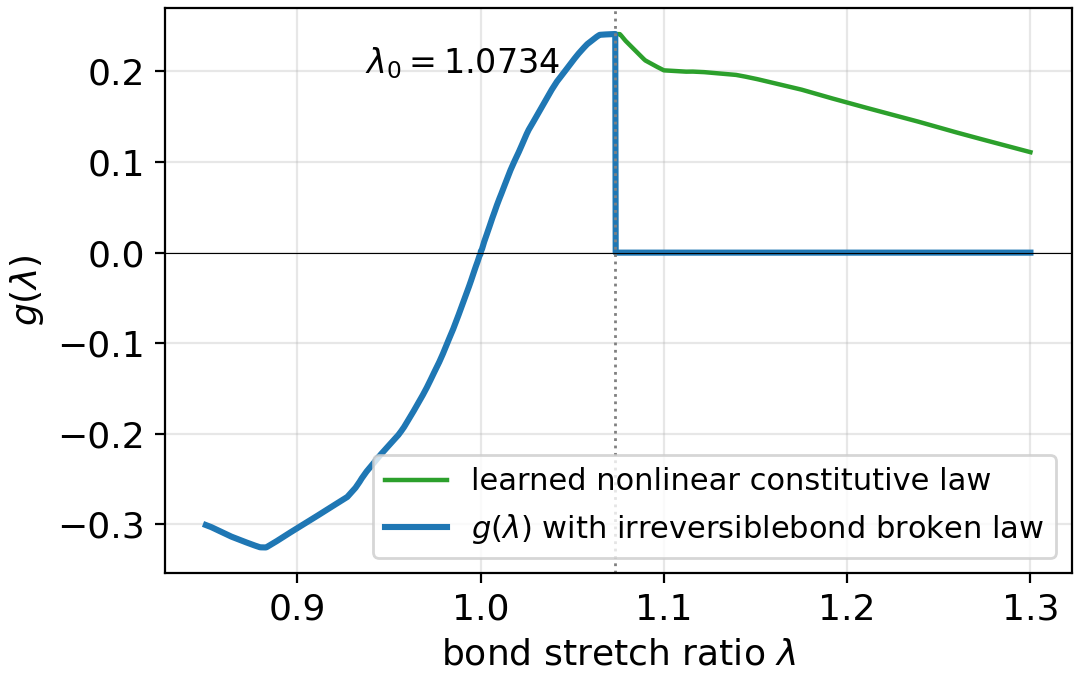}
\includegraphics[width=0.42\linewidth]{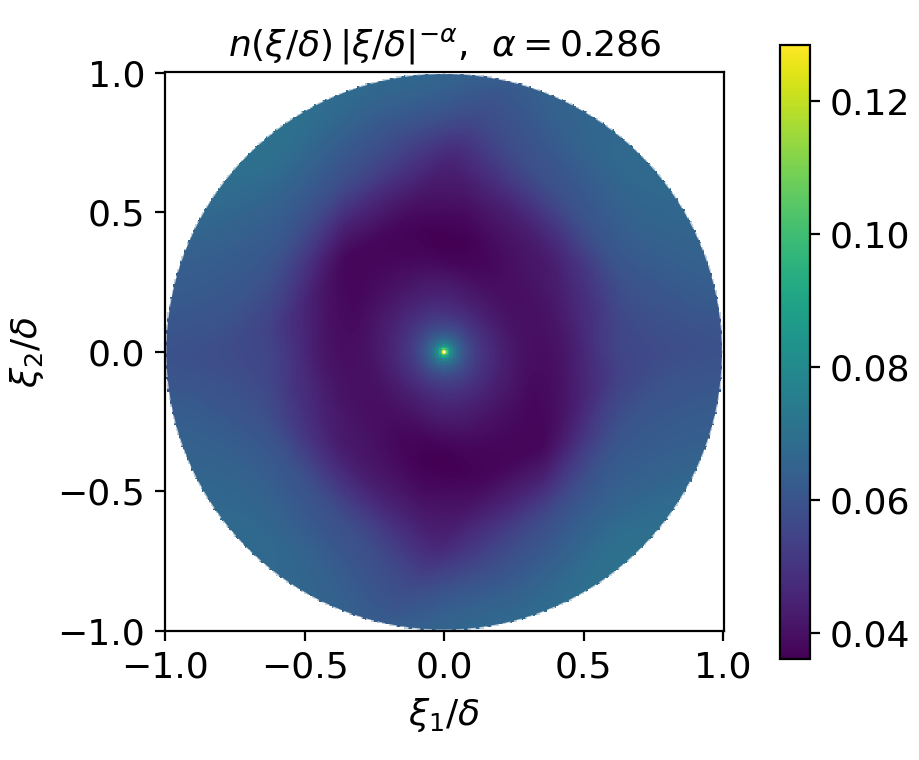}
\caption{The learned MNO constitutive functions.  Left: learned nonlinear constitutive law (green) and the downstream $g(\lambda)$ after the irreversible bond-breaking rule is applied (blue).  Right: learned kernel $n(\xib/\delta)\,|\xib/\delta|^{-\alpha}$. }
\label{fig:lr3e4-kernel}
\end{figure}

Since the MD simulations are quasi-static, for the purpose of training we also adopt a static MNO model. Denoting the three sets of coarse-grained MD data as $\mcD^{MD}_{train/valid/test}=\{\ub^{(i)}_{train/valid/test}(\xb),\bb^{(i)}_{train/valid/test}(\xb)\}$, the training/validation/test error is defined as:
\begin{equation}\label{eq:MD-force-metrics}
\begin{split}
&E_{train/valid/test}^{MD}
:=\left[
\frac{1}{\#(\mcD^{MD}_{train/valid/test})}\sum_{(\ub,\bb)\in \mcD^{MD}_{train/valid/test}}\dfrac{\vertii{\mcF_{\delta,\Delta x}^{MNO,2D,NN}[\ub;\theta]+\bb}_{l^2}}{\vertii{\bb}_{l^2}}
\right]^{1/2}.
\end{split}
\end{equation}
Following the practice in Section~\ref{subsec:synthetic-data}, we
parameterize the constitutive functions as
\begin{equation}
\begin{aligned}
g_\theta(\lambda)&=\operatorname{sign}(\lambda-1)\,
  \operatorname{softplus}\!\left(N_g(\lambda-1)\right),\\
n_\theta(\xib/\delta)&=N_n(\xib/\delta),\\
\alpha_\theta&=\operatorname{softplus}(a)+10^{-6},
\end{aligned}
\label{eq:md-network-parameterization}
\end{equation}
where $N_g$ and $N_n$ are fully connected neural networks with four
hidden layers of width $128$ and ReLU activations ($99{,}971$
trainable parameters in total), and $a$ is a trainable scalar
initialized such that $\alpha_\theta \approx 0.913$.  This
parameterization enforces $g_\theta(1)=0$, the non-negativity
constraint $g_\theta(\lambda)\geq 0$ for $\lambda \geq 1$ required by
the rescaling law of Section~\ref{sec:rescale_s0}, and
$\alpha_\theta>0$.  In contrast to the isotropic setting of
Section~\ref{sec:synthetic-verification}, the kernel network here
takes the full two-dimensional normalized bond vector $\xib/\delta$
as input, so no isotropy is imposed a priori.  The horizon is treated
as a hyperparameter and fixed at $\delta = 3.01\Delta x = 15.05\,lu$.
The training loss follows \eqref{eq:loss}, with the force residual of
\eqref{eq:nonlocElasticity_comp_2d} evaluated on the interior
$15\times 15$ nodes of each coarse-grained sample (excluding a
boundary collar of width $\delta$, where the coarse-graining
aggregation is incomplete).  The model is trained using AdamW, with learning rates $1.5\times 10^{-4}$ for the networks and $2\times 10^{-3}$ for $\alpha_\theta$ and weight decay $5\times 10^{-4}$. The learned model converges to the fractional exponent $\alpha = 0.286$, with the learned nonlinear constitutive law and kernel demonstrated in Figure \ref{fig:lr3e4-kernel}.  
In the learned nonlinear constitutive law, the tension-side peak occurs at $\lambda_0 = 1.073$: its rising branch matches that of the underlying atomic pair force, but the peak indicates a slightly earlier fracture criterion compared with the critical bond stretch ratio of $1.15$ in the MD simulations. Thus, in the MNO model,
bond breakage can occur at a lower strain than the peak strain in the underlying MD model.
This difference arises because the MNO model has a larger length scale than the MD lattice: the MNO grid size ($5\,lu$) is larger than the MD lattice spacing ($1.40\,lu$).
If a large bond strain occurs between two atoms in the MD model under a given loading function, the
longer MNO bonds in the same direction may not experience the same large strain.
However, if failure occurs in the MD model, the MNO model is trained to fail at the same time.
Thus, the bond strains at failure in the MNO model are slightly less than the MD peak bond strain.
This observation is also consistent with Lemma \ref{lemma:2}, which anticipates that the critical stretch of mesoscale models decreases as the internal length scale increases. With the irreversible fracture assumption, for downstream evaluations and predictions we irreversibly break a bond once its
stretch ratio exceeds $\lambda_0$. The resulting $g(\lambda)$ is illustrated by the blue line in the left plot of Figure \ref{fig:lr3e4-kernel}. The right plot of Figure \ref{fig:lr3e4-kernel} shows that the learned kernel is not isotropic, reflecting the hexagonal lattice microstructure of the MD system.

To quantitatively evaluate the learned model, we again consider both prediction errors and operator errors, following the practice for the synthetic datasets in Section \ref{subsec:synthetic-evaluation}. While the former provides a direct evaluation of the model performance in new and unseen loading scenarios, the latter aims to validate the learned constitutive law. Since the ground-truth constitutive law is not available here, instead of measuring function errors we evaluate the constitutive-law errors by comparing two macroscopic quantities measured from MD simulations against their counterparts computed from the learned MNO model.

\textbf{Macroscopic quantity 1: Young's modulus and strain energy density.} 
To determine the Young's modulus,
we consider an MD grid of dimensions $100\,lu\times100\,lu$ and thickness $H=0.935\,lu$,
hence cross-sectional area in the plane $x_2=0$ is $A = 93.5\,lu^2$.
The mesh is deformed statically such that the strain tensor is
\begin{equation}\label{eqn:strain}
  \boldsymbol{\epsilon}
  =
  \begin{pmatrix}
    -0.0033 & 0 \\
     0      & 0.01
  \end{pmatrix}.
\end{equation}
The MD calculation is run for one time step, and the total force through the
cross-sectional plane is computed.
This total force is divided by the cross-sectional area $A$ to obtain the stress tensor.
For the MD interatomic potential described above, this turns out to be
\[
  \boldsymbol{\sigma}
  =
  \begin{pmatrix}
    0 & 0 \\
    0 & 3.82\,su
  \end{pmatrix}
  =
  \begin{pmatrix}
    0 & 0 \\
    0 & 0.0382\,GPa
  \end{pmatrix}.
\]

\begin{figure}[h]
  \centering
  \includegraphics[width=0.48\linewidth]{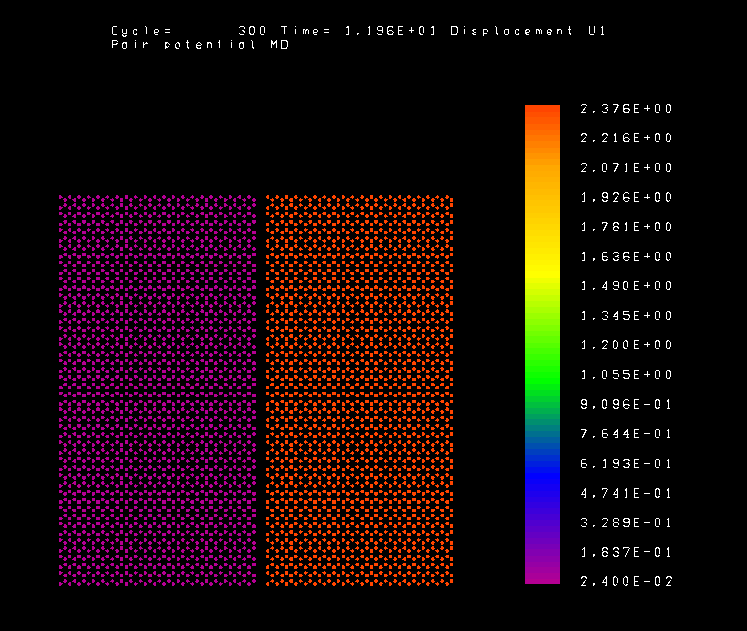}
  \hfill
  \includegraphics[width=0.48\linewidth]{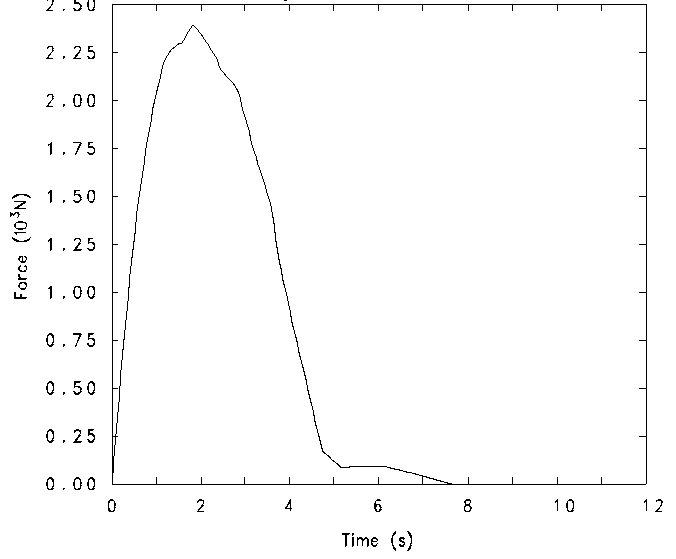}
  \caption{MD grid (left: displacement field $U_1$) and total force vs.\ time (right)
    for the critical energy release rate for fracture calculation.}
  \label{fig:md_grid_force}
\end{figure}

Because this is a uniaxial state of stress (only the $(2,2)$ component is nonzero), the
Young's modulus is
\[
  E = \frac{\sigma_{22}}{\epsilon_{22}}
    = 382{su}
    = 3.82GPa.
\]
With the Young's modulus, we can then evaluate the strain energy density corresponding to the prescribed strain tensor $\boldsymbol{\epsilon}$ in \eqref{eqn:strain}. Following from linear elasticity $W = \frac12
\boldsymbol{\sigma}:\boldsymbol{\epsilon}$ and noting that the prescribed
strain state is precisely a uniaxial stress state, the strain
energy density of the MD system is therefore
\begin{equation}\label{eq:W_MD}
  W^{MD} = \tfrac12\,\sigma_{22}\,\epsilon_{22}
         = \tfrac12\,E\,\epsilon_{22}^2
         = \tfrac12 \times 382\,su \times (0.01)^2
         = 1.91\times 10^{-2}\,su
         = 1.91\times 10^{5}\ kg/m/s^2,
\end{equation}
which serves as the reference value for validating the learned model.  This quadratic expression is valid in the small-strain linear regime probed here ($1\%$ strain). At larger strains both the MD response and the learned model depart from it through the nonlinearity of the bond force.

To evaluate the corresponding strain energy density from MNO, from the 2D MNO formulation \eqref{eq:nonlocElasticity_comp_2d} we have:
\begin{align}
\nonumber W^{MNO,2D}_\delta(\xb,t) = &\frac12 \int_0^H \int_{B^{2D}_\delta (0)} w_\delta^{MNO,2D} (\lambda[\ub],\xib,t) d\xib dh \\
=&\frac12 H \int_{{B^{2D}_\delta (0)}}c_0 \delta^{-d} n(\xib/\delta) |\xib/\delta|^{-\alpha+1} \beta(\lambda[\ub](\xib,\xb,t)) \,d\xib,\label{eq:W_2d}
\end{align}
with unit $eu/lu^3=mu/lu/tu^2$. Here the integral in $\xib$ is taken over a 2D disk. Numerically, we consider the same affine deformation, $\etab = \boldsymbol{\epsilon}\,\xib$ with
$\boldsymbol{\epsilon} = \mathrm{diag}(-0.0033,\, 0.01)$, for which
the bond stretch reduces to $\lambda(\xib) =
|(\boldsymbol{I}+\boldsymbol{\epsilon})\xib|/|\xib|$. Then, the
antiderivative $\beta(\lambda) = \int_1^\lambda g_\theta(s)\,ds$ is
computed by trapezoidal integration on a fine grid
($\Delta\lambda = 10^{-5}$), and the disk
integral is evaluated with the same Riemann quadrature and bond
stencil used in training ($\delta/\Delta x = 3.01$,
cf.~\eqref{eqn:Riemann_sum}).


\textbf{Macroscopic quantity 2: critical energy release rate for fracture.} 
This is the amount of energy, per unit of crack growth area, consumed by a 
crack as it grows.
To evaluate this from MD, a rectangular grid is initialized with zero strain and
displacement.
As shown in Figure \ref{fig:md_grid_force}, the horizontal midplane $x_2=0$ is designated as the crack plane.
With each half prescribed to move as a rigid body, the two are slowly separated
across this plane until the particles on either side no longer interact.
The integral of the total force $P(u)$ across the plane with respect to the separation
distance $u$ is the total work required to separate the two halves.
Dividing this by the cross-sectional area $A$ results in an estimate for the
critical energy release rate $G$:
\[
  G = \frac{1}{A}\int_0^{\infty} P(u)\,du.
\]
The numerical result is $G = 16.4\,eu/lu^2$.
Converting to SI yields:
\begin{equation}
G^{MD}
  = 16.4\,\frac{eu}{lu^2}
  \cdot 10^{-23}\,\frac{\mathrm{J}}{eu}
  \cdot 10^{20}\,\frac{lu^2}{\mathrm{m}^2}
  = 0.0164\,kg/s^2.
\end{equation}
This is highly idealized because it does not take into account energy dissipation 
due to deformation in the process zone of a crack.
However, for a brittle material such as this, it provides a good estimate of the critical energy release rate.

Correspondingly, the fracture energy from MNO can be evaluated as:
\begin{align}
\nonumber G^{MNO,2D}_{{\delta}}(\lambda_0(\delta))=&\int_0^H \int_{0}^{{\delta}}\int_{H^{2D}_{{\delta}}(z)}  w_\delta^{MNO,2D}(\lambda_0(\delta),\xib)\, d\xib \, dz\, dh\\
\nonumber=&c_0 {\delta}^{-d} H\int_{0}^{{\delta}}\int_{H^{2D}_{{\delta}}(z)}  n(\xib/{\delta}) |\xib/{\delta}|^{-\alpha+1} \beta(\lambda_0(\delta))\, d\xib \, dz\\
=&c_0 H \delta\, \beta(\lambda_0)
  \underset{\zeta_2>0}{\int_{B^{2D}_1(0)}} \zeta_2\, n(\zetab)\,
  |\zetab|^{-\alpha+1}\, d\zetab,\label{eq:G_2d}
\end{align}
with unit $eu/lu^2=mu/tu^2$. Herein, the inner integral is taken over a 2D cap, and the last equality is obtained by substituting $\varepsilon = z/\delta$ and swapping the order of the double integral. Numerically, the critical stretch is obtained from the learned bond law: $\lambda_0 = \mathrm{argmax}_{\lambda > 1}\,
g_\theta(\lambda) = 1.073$, and the associated bond-breaking
energy $\beta(\lambda_0) = \int_1^{\lambda_0} g_\theta(s)\,ds$ is
computed by trapezoidal integration.  The half-disk integral is evaluated with the
same training-stencil quadrature as in the evaluation for $W^{MNO,2D}$.

\textbf{Quantitative evaluations of the learned MNO. } In Table \ref{tab:md} part (a) we report the prediction errors of MNO on three coarse-grained MD datasets, and in part (b) we quantify the learned constitutive law in MNO through the two macroscopic quantities. One can see that in prediction tasks the MNO errors are below $10\%$, while the underlying constitutive law matches well with MD on both the strain energy density and the critical energy release rate (with $2.88\%$ and $2.01\%$ errors, respectively).

\begin{table}[H]
\centering
\renewcommand{\arraystretch}{1.21}
\caption{Learning results from MD.}
\label{tab:md}
\setlength{\tabcolsep}{3pt}
\begin{tabular}{@{}l|cccccc@{}}
\toprule
\multicolumn{7}{l}{(a) Learnt Parameters and Prediction Errors} \\
\midrule
\addlinespace[3pt]
Model & $\delta$ & $\alpha$ & $\lambda_0$ & Training Error &  Validation Error &  Test Error\\
\midrule
MNO & $15.05\,lu$    & $0.286$ & $1.073$ & $9.45\%$ & $8.30\%$ & $9.44\%$\\
\bottomrule
\end{tabular}

\medskip
\begin{tabular}{@{}l|cccc@{}}
\toprule
\multicolumn{5}{l}{(b) Macroscopic Quantities and Errors} \\
\midrule
\addlinespace[3pt]
Model & Strain Energy Density $W$ & Error of $W$& Critical Energy Release Rate $G$ & Error of $G$ \\
\midrule
MNO & $1.97\times 10^5\,kg/m/s^2$ & $2.88\%$ & $0.0167\, kg/s^2$ & $2.01\%$\\
MD & $1.91\times 10^5\,kg/m/s^2$ & N/A & $0.0164\, kg/s^2$ & N/A\\
\bottomrule
\end{tabular}
\end{table}

\subsection{Validation on Quasi-Static Simulations}\label{sec:validation}

\begin{figure}[htbp]
  \centering
  \includegraphics[width=0.2\textwidth]{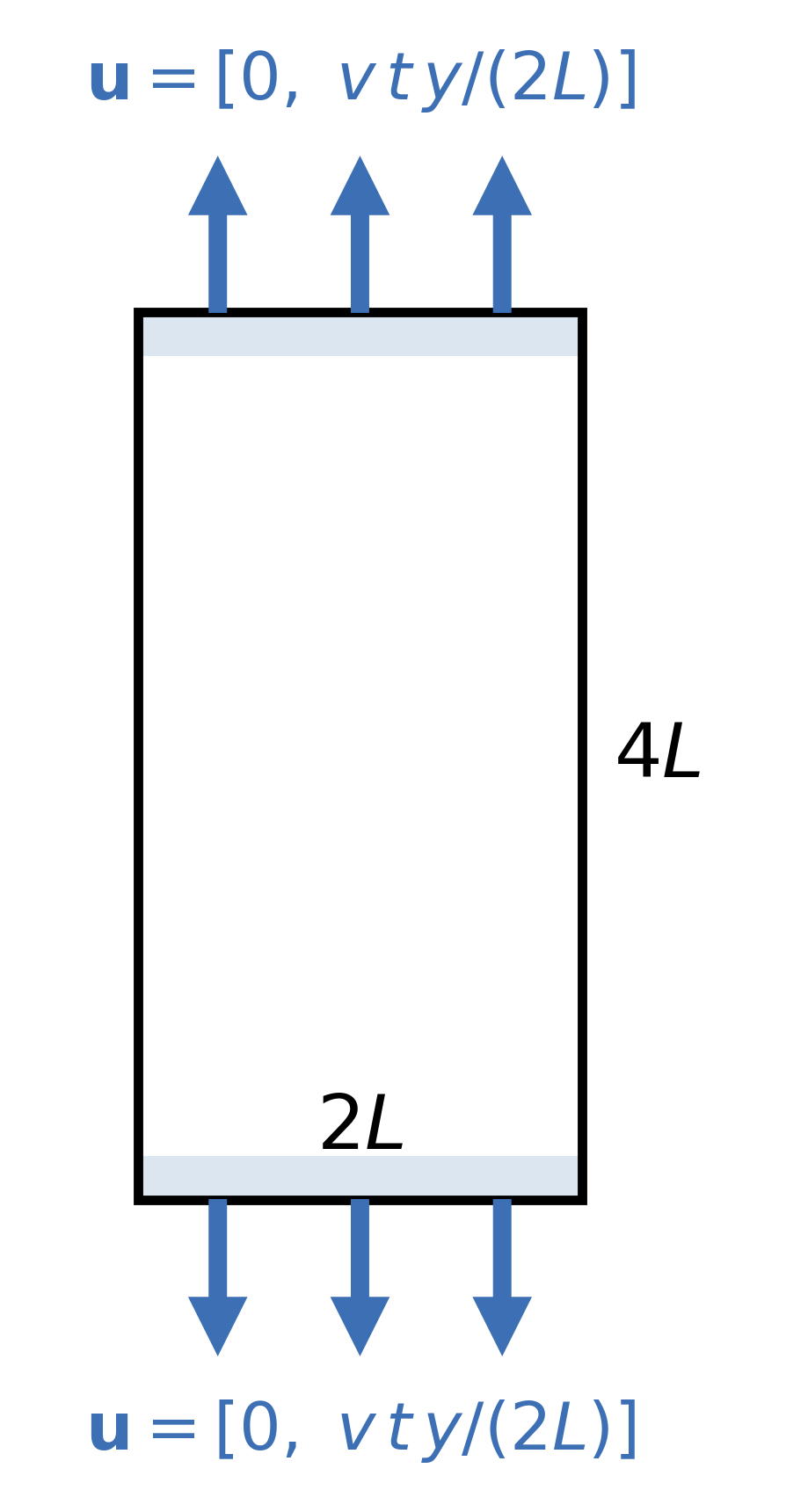}
  \includegraphics[width=0.2\textwidth]{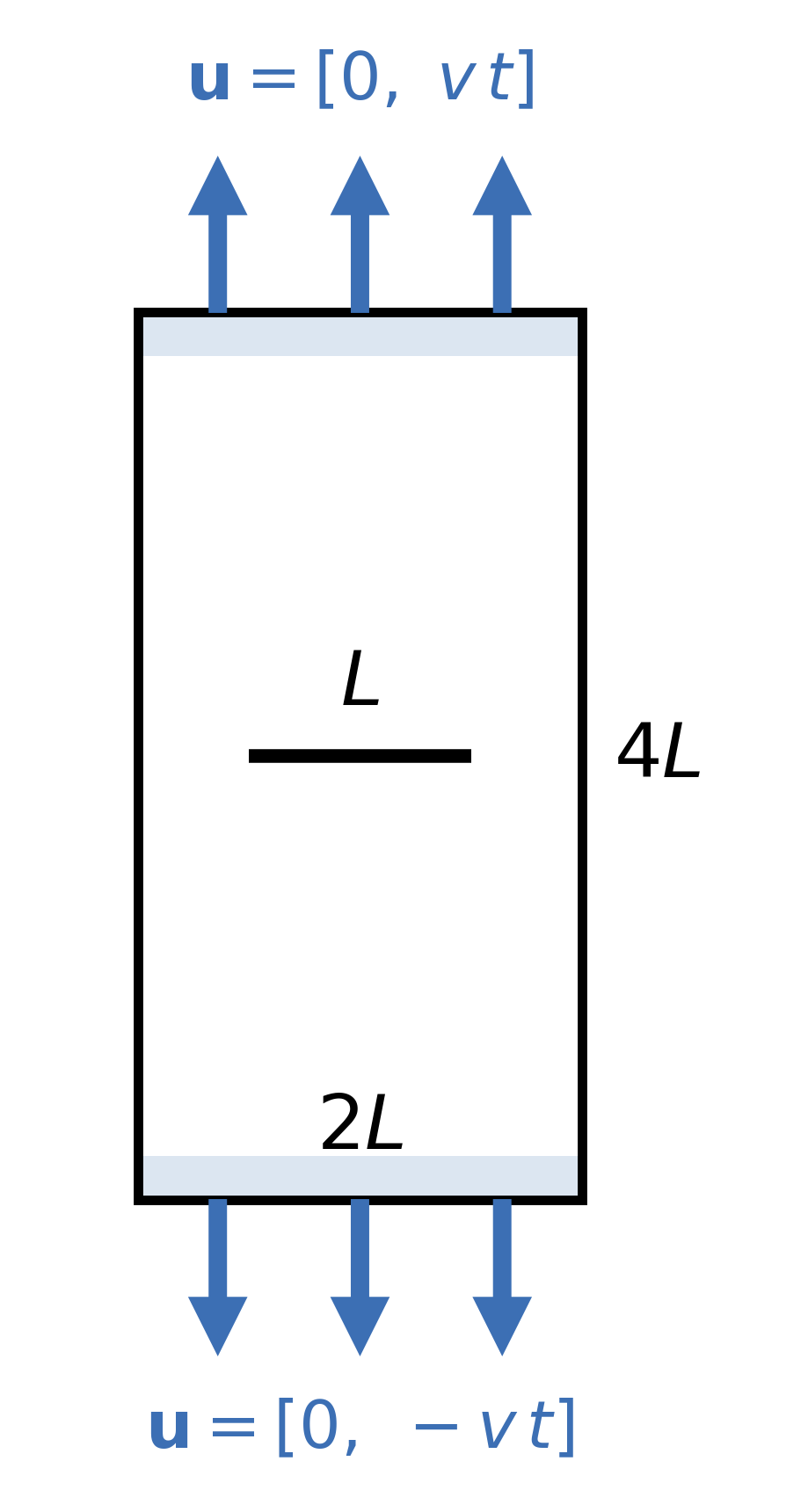}  \includegraphics[width=0.58\textwidth]{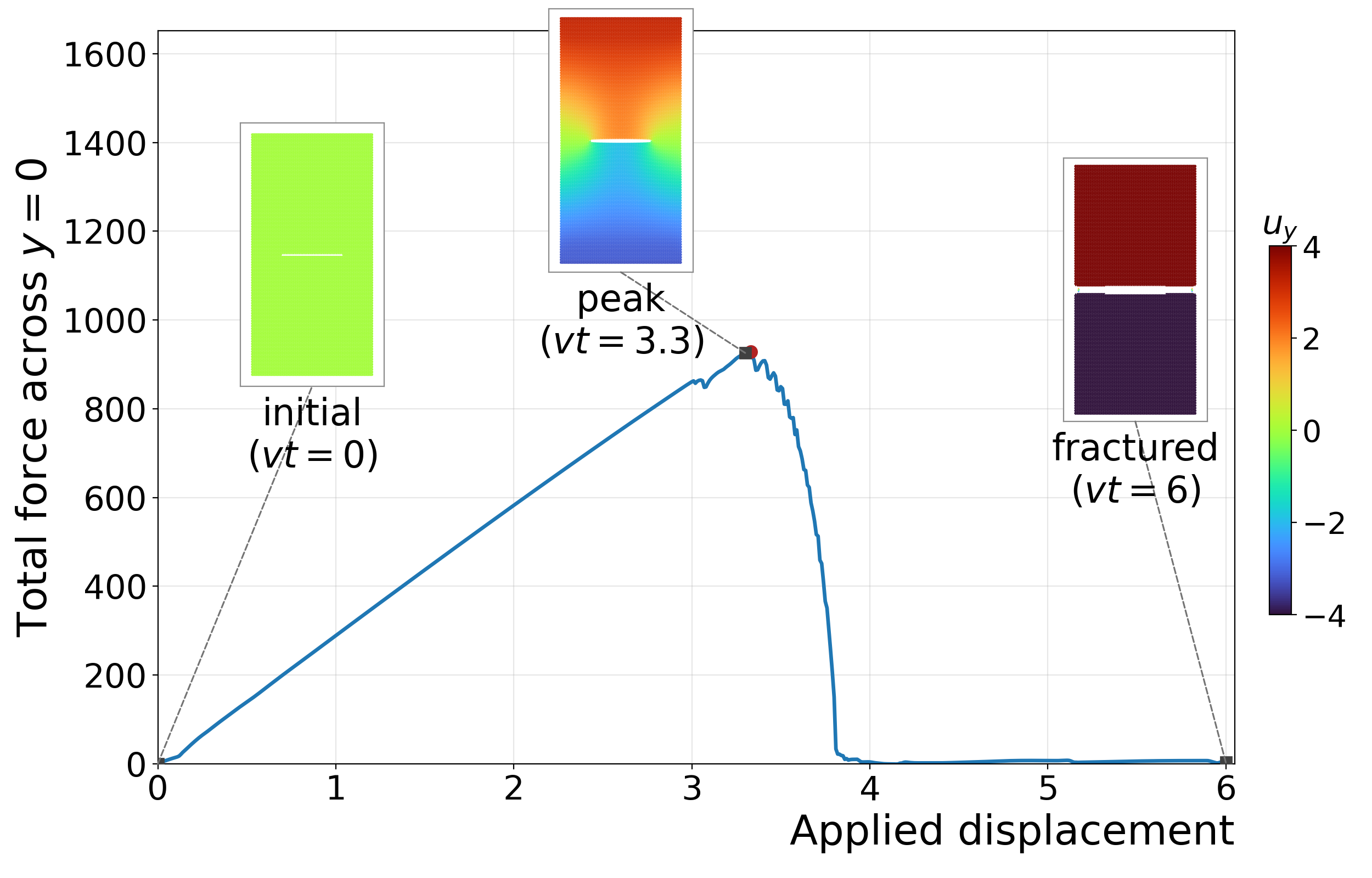}
  \caption{Settings of the validation tests: a $2L \times 4L$ plate loaded in tension by prescribing the vertical displacement
  $\ub = [0, \pm v\,t]$ on strips of width $10\,lu$
  (shaded) along the top and bottom boundaries.  Left: plate without pre-notched crack.  Middle: plate with a horizontal
  pre-notched crack of length $L$ centered on the midplane $y=0$. Right: a demonstration of the anticipated behavior in the pre-notched plate, showing the evolution of total forces and displacement fields in MD simulations when $L=50\,lu$.}
  \label{fig:validation-setup}
\end{figure}

We now validate the learned model on displacement-controlled tension tests that lie well outside the training distribution, against MD simulations of the same settings. As shown in Figure \ref{fig:validation-setup}, we consider $2L \times 4L$ plates subject to Dirichlet boundary conditions at the top and bottom boundaries. With a prescribed pull rate $v$, we quasi-statically increase the vertical displacements on strips of width $10\,lu$ as:
\begin{equation}
\ub_{BC}(x,y,t)=\left\{\begin{array}{cl}
{[0,\ v\,t]},  & y\in[2L-10,\ 2L],\\
{[0,\ -v\,t]}, & y\in[-2L,\ -2L+10].
\end{array}\right.
\end{equation}
Denoting the union of boundary strips as $\partial\Omega:=[-L,L]\times ([-2L,\ -2L+10]\bigcup [2L-10,\ 2L])$ and the rest of the plate as $\Omega:=[-L,L]\times [-2L+10,\ 2L-10]$, at each time instance, we solve for the displacement field following:
\begin{align}\label{eqn:MNO_quasi}
\mcF^{MNO,2D}_{\tilde{\delta}}[\ub](x,y,t) = \mathbf{0}, &\text{ for }(x,y)\in \Omega,\\
\ub(x,y,t) = \ub_{BC}(x,y,t), &\text{ for }(x,y)\in \partial\Omega,
\end{align}
where $\mcF^{MNO,2D}_{\tilde{\delta}}$ is the MNO model trained from coarse-grained MD data and rescaled from $\delta$ to $\tilde{\delta}$. The primary quantity of comparison is the total force transmitted across the midplane, obtained by summing the bond forces of all bonds that cross $y=0$:
\begin{equation}\label{eqn:total_force}
\begin{aligned}
    F(t):=& \int_0^H\int_{\xb\in\Omega\bigcap \real^2_{+}}\int_0^H\int_{\zb-\xb \in B^{2D}_{\tilde{\delta}} (\xb),\, \zb\in \Omega\bigcap \real^2_{-}}  \fb^{MNO,2D}_{\tilde{\delta}}(\etab,\zb-\xb,t)\cdot \eb_2 d\zb dh' d\xb dh\\
    \approx & H^2 {\sum_{\xb_i\in \chi\cap \mathbb{R}^2_+}}\;
    {\sum_{\xb_j\in B_{\tilde{\delta}}(\xb_i)\cap\chi\cap \mathbb{R}^2_-}}
    \fb^{MNO,2D}_{\tilde{\delta}}(\etab_{i,j-i},\xib_{i,j-i},t)\cdot \eb_2\, \Delta x^4.
\end{aligned}
\end{equation}
Here, $\eb_2:=[0,1]$, $\mathbb{R}^2_+ := \{(x,y)\in\mathbb{R}^2 : y > 0\}$, and $\mathbb{R}^2_- := \{(x,y)\in\mathbb{R}^2 : y < 0\}$. The total force trajectory is compared against the corresponding reaction force across the $y=0$ plane measured in MD simulations. In addition, we also compare the predicted displacement fields against the ones from MD simulations.

\begin{table}[H]
\centering
\caption{Errors of the learned model against MD in the validation
tests, for the peak total force as defined in \eqref{eqn:total_force} and the displacement field at both peak and final time instances. For the non-fracture case the total force is monotonically increasing, so we report the peak force at the final time instance $vt=5\,lu$ and no peak time is reported.}
\label{tab:disp_errors}
\setlength{\tabcolsep}{3.5pt}
\begin{tabular}{lcccccc}
\toprule
\multicolumn{3}{c}{Setting} & \multicolumn{4}{c}{Relative Error}\\
\cmidrule(lr){1-3}\cmidrule(lr){4-7}
test case & domain & $\tilde{\delta}$ & peak force & peak time
 & peak displacement& final displacement\\
\midrule
Non-fracture              & $100\times 200$ & $15.05$ & $1.61\%$ & ---
 & $1.32\%$ & ---\\
Pre-notched, Setting 1      & $100\times 200$ & $15.05$ & $3.98\%$
 & $0.03\%$ & $2.97\%$ & $11.48\%$\\
Pre-notched, Setting 2     & $200\times 400$ & $15.05$ & $4.83\%$
 & $8.12\%$ & $3.84\%$ & $9.43\%$\\
Pre-notched, Setting 3 & $200\times 400$ & $30.10$ & $5.29\%$
 & $0.61\%$ & $4.00\%$ & $15.79\%$\\
\bottomrule
\end{tabular}
\end{table}

\begin{figure}[h]
  \centering
\includegraphics[width=0.48\linewidth]{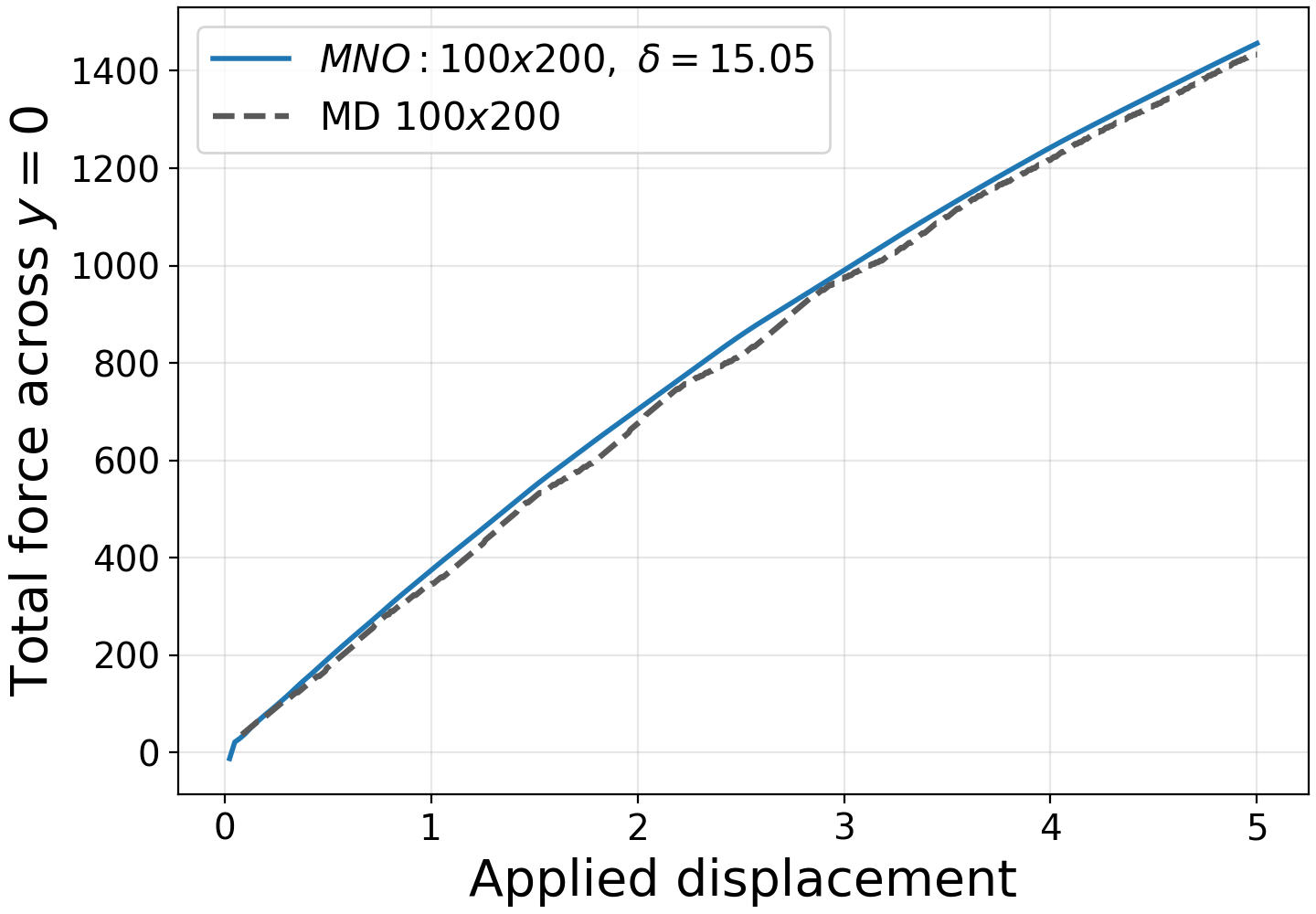}
\includegraphics[width=0.48\linewidth]{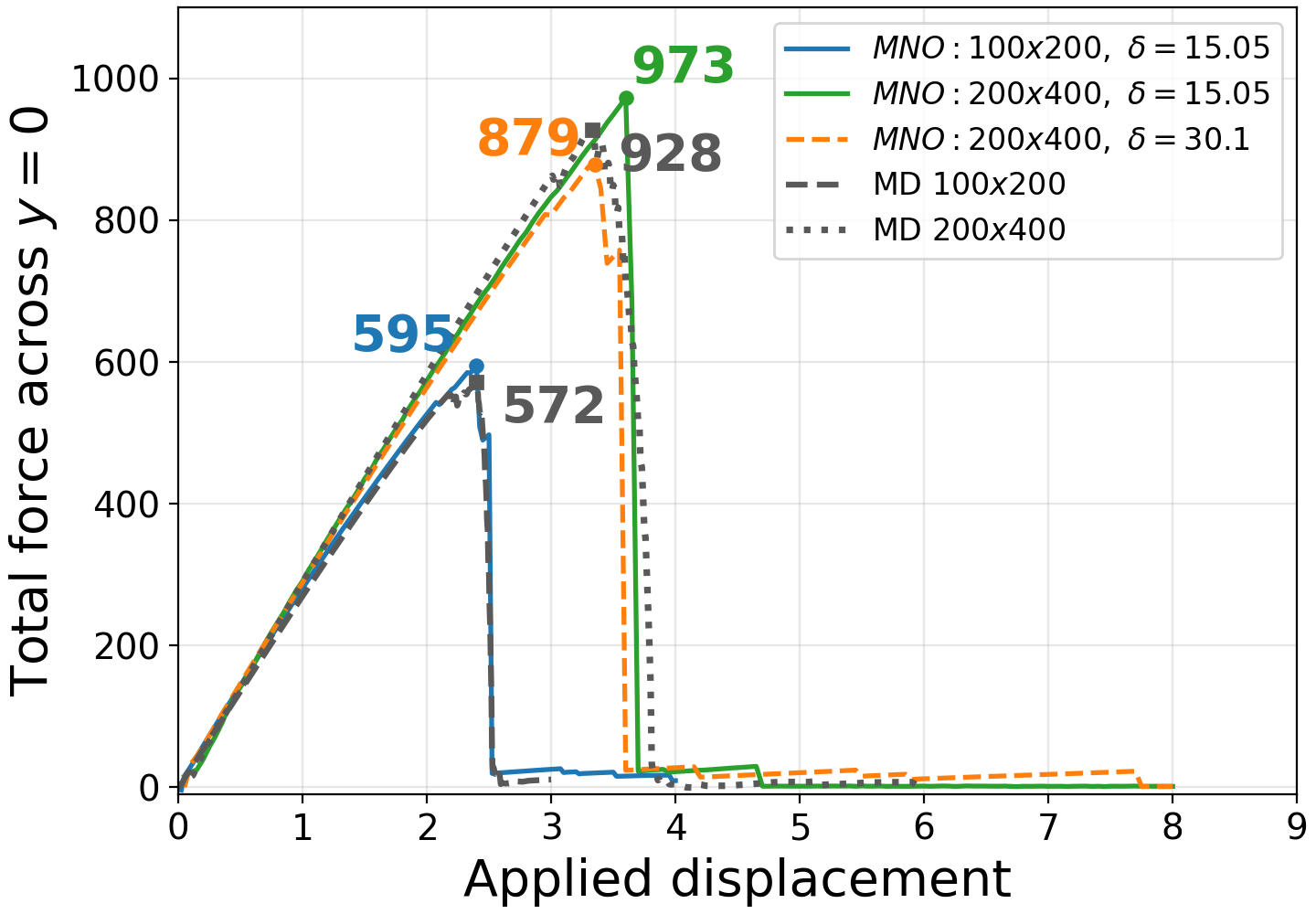}
  \caption{Comparison of total forces from MNO against MD in the validation tests. Left: results in the non-fracture plate case. Right, results in the pre-notched fracture case.}
  \label{fig:totalforce}
\end{figure}

To validate both the learned nonlinear constitutive law and the fracture rule, we apply the MNO model in two domain settings:
\begin{itemize}
\item \textbf{Plate without fracture.} As shown in the left plot of Figure \ref{fig:validation-setup}, we first consider a plate without fracture. Within this setting, we apply the learned constitutive model with $L=50\,lu$ and $v = 0.025\,lu/tu$, using the learned MNO model with original $\Delta x=5\,lu$ and ${\tilde{\delta}}=3.01\Delta x$ as in Section \ref{section:MDlearn}. In this example, the total force is anticipated to increase nonlinearly with the increase of applied displacement. This study aims to validate the nonlinear relationship between deformation and total force.
\item \textbf{Plate with pre-notched fracture.} As demonstrated in the middle plot of Figure \ref{fig:validation-setup}, a plate containing a horizontal pre-notch of length $L$ centered on the midplane $y=0$ is considered. In the MNO simulation, the pre-notch is realized by removing all bonds that cross the notch segment. To further validate the transferability of the MNO at a larger scale, within this setting we apply the learned constitutive model with three geometry/horizon settings:
\begin{enumerate}
\item{\textit{Setting 1 (small internal length, small domain).}} Original $\Delta x=5\,lu$ and ${\tilde{\delta}}=3.01\Delta x$, on a small domain with $L=50\,lu$ and $v = 0.025\,lu/tu$.
\item{\textit{Setting 2 (small internal length, large domain).}} Original $\Delta x=5\,lu$ and ${\tilde{\delta}}=3.01\Delta x$, on a larger domain with $L=100\,lu$ and $v = 0.05\,lu/tu$.
\item{\textit{Setting 3 (large internal length, large domain).}} Larger $\Delta x=10\,lu$ and ${\tilde{\delta}}=3.01\Delta x$, on a larger domain with $L=100\,lu$ and $v = 0.05\,lu/tu$.
\end{enumerate}
As shown in the right plot of Figure \ref{fig:validation-setup}, as the total force increases with the applied displacement, the response is at first linearly elastic.
As the loading increases, bonds around the notch tips experience stretches that approach the critical ratio $\lambda_0$. Upon further loading, some bonds at the notch tips exceed $\lambda_0$ and break irreversibly. Each broken bond transfers its load to its neighbors, driving them past the critical stretch in turn. Because this is an unstable crack configuration, the cracks propagate rapidly within a few loading steps. Once the plate is fully separated into two halves, the interior strain drops toward zero, and the transmitted force settles to a small residual. By comparing the simulated total force ($F(t)$) and displacement fields at the peak against the corresponding MD simulations, we aim to validate the learned fracture rule in the MNO. Additionally, as the domain size doubles, the maximum force is also anticipated to increase, which should result in an increase of the peak force from Setting 1 to Setting 2. Lastly, to validate our mesoscale constitutive law across different scales, we compare the simulation results from Settings 2 and 3. With the same domain and boundary condition settings, these two settings are anticipated to capture a consistent material behavior across different internal lengths ${\tilde{\delta}}$.
\end{itemize}


\begin{figure}[t!]
  \centering
  \includegraphics[width=0.6\linewidth]{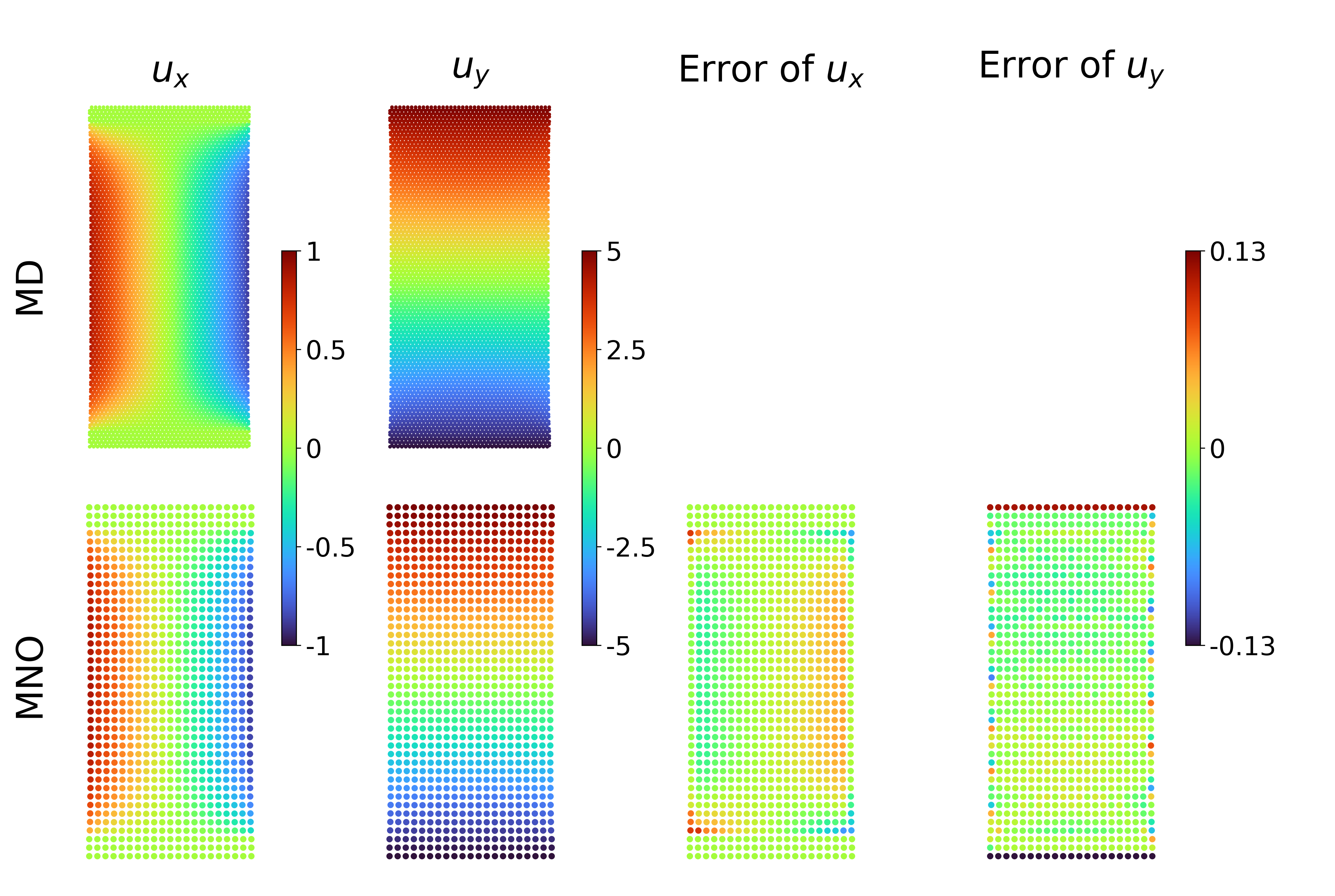}
  \caption{Comparison of final displacement from MNO against MD in the validation test on non-fracture plate, with a relative $L^2(\Omega)$ error of $1.32\%$.}
  \label{fig:particles-nofracture}
\end{figure}

\begin{figure}[t!]
  \centering
  \includegraphics[width=0.8\linewidth]{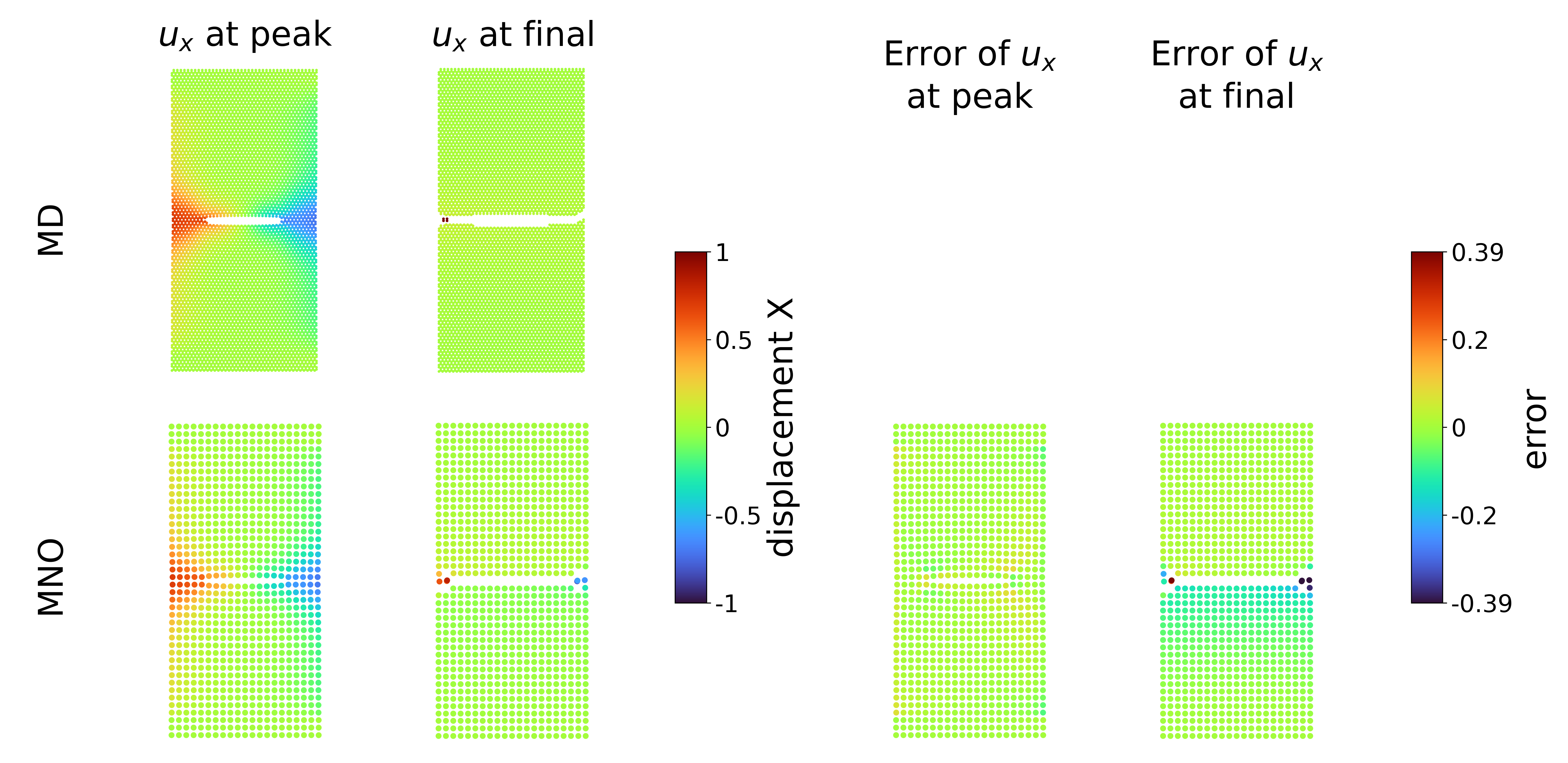}
  \includegraphics[width=0.8\linewidth]{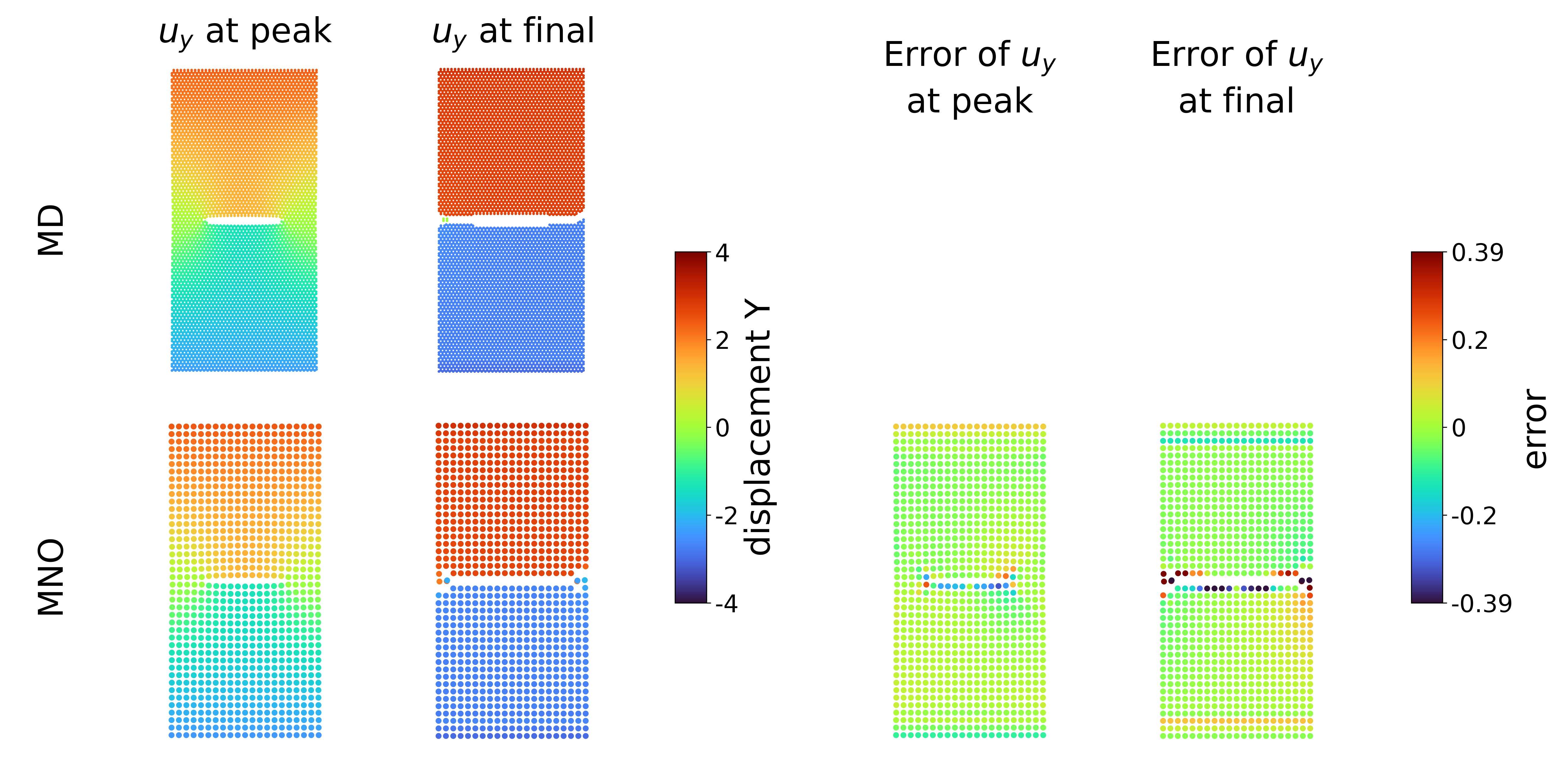}
  \caption{Comparison of peak and final displacement fields from MNO Setting 1 against MD in the validation test on a $100\times 200$ pre-notched plate, with relative $L^2(\Omega)$ errors of $2.97\%$ at peak time and $11.48\%$ at final time.}
  \label{fig:particles-L50}
\end{figure}

\begin{figure}[t!]
  \centering
  \includegraphics[width=0.8\linewidth]{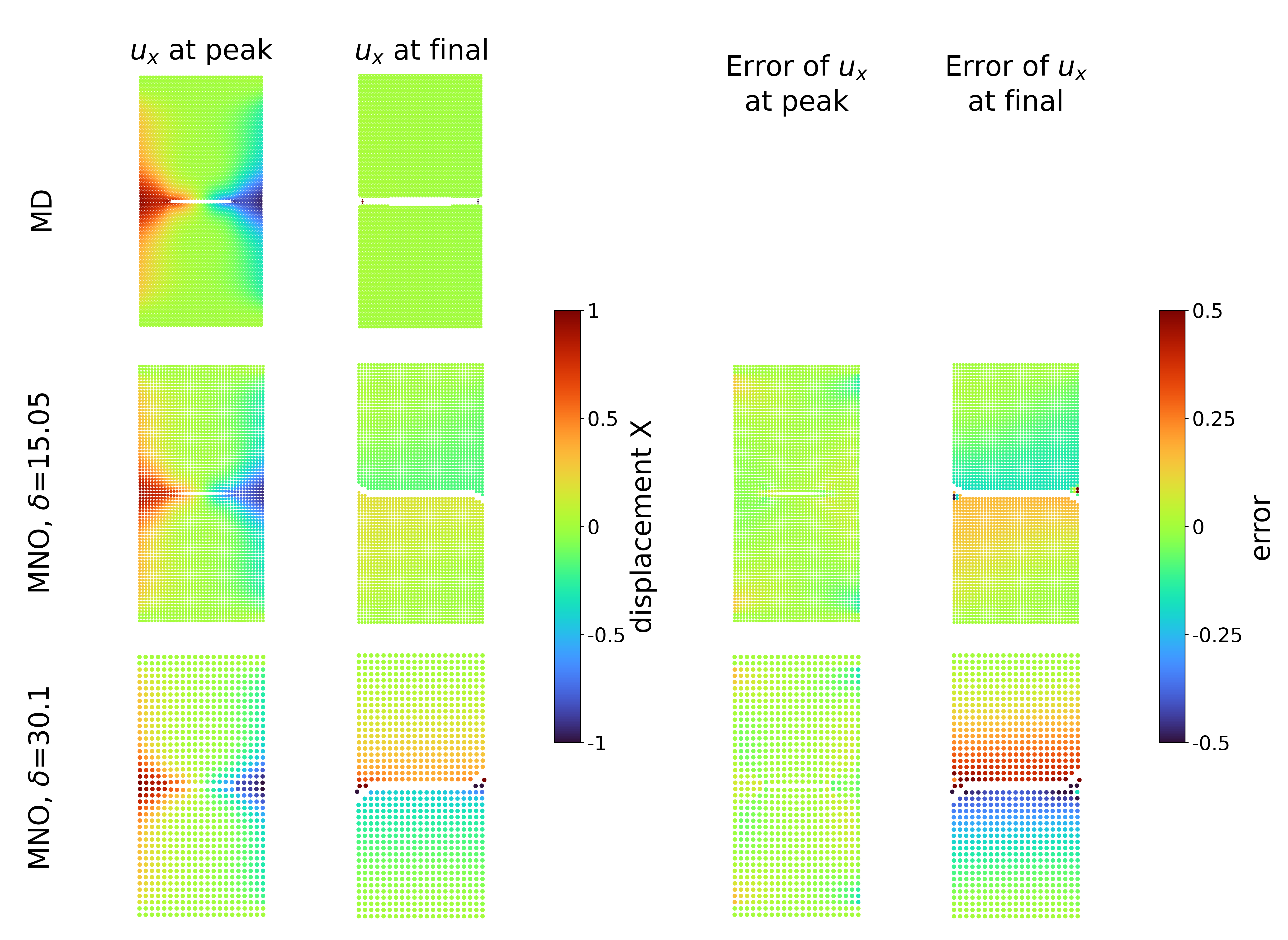}
  \includegraphics[width=0.8\linewidth]{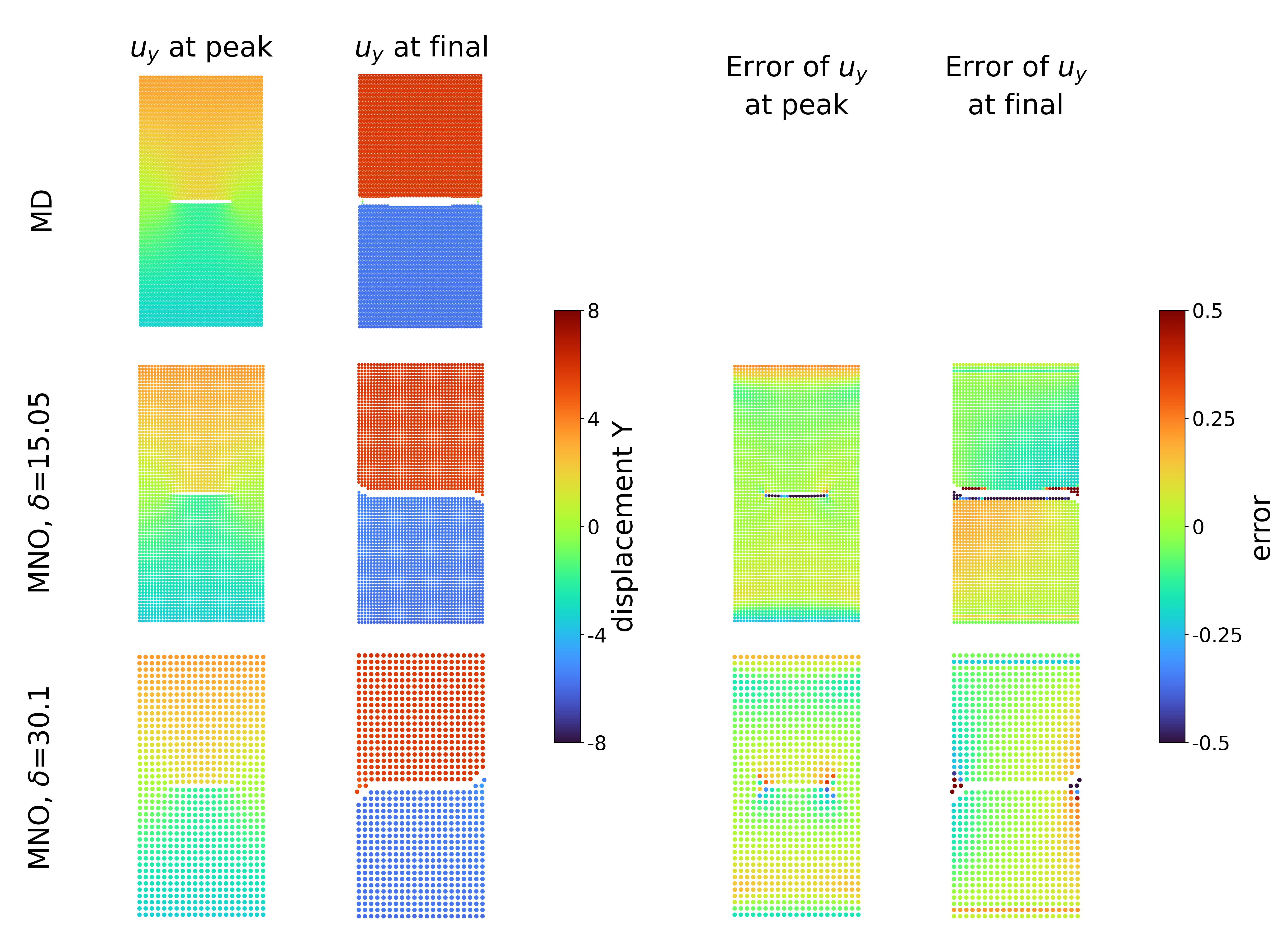}
  \caption{Comparison of peak and final displacement fields from MNO Settings 2 and 3 against MD in the validation test on a $200\times 400$ pre-notched plate. For the MNO with the original $\tilde{\delta}=15.05\,lu$ (Setting 2), the relative $L^2(\Omega)$ errors are $3.84\%$ at peak time and $9.43\%$ at final time. For the MNO with a rescaled ${\tilde{\delta}}=30.10\,lu$ (Setting 3), the relative $L^2(\Omega)$ errors are $4.00\%$ at peak time and $15.79\%$ at final time.}
  \label{fig:particles-L100}
\end{figure}

\textbf{Comparison on the total force trajectory.} We first plot the results of total transmitted force $F(t)$ as functions of the applied displacement in Figure~\ref{fig:totalforce}, and report the errors of peak force and peak time in Table \ref{tab:disp_errors}. One can see that the total force is monotonically increasing in the non-fracture case, and hence we only validate the peak force at the final time instance. In the non-fracture case, the learned MNO model reproduces the measured MD force with relative errors of $2.74\%$ over the whole loading history and $1.61\%$ at the final time instance, confirming that the learned nonlinearity is consistent with MD. In the pre-notched case, the total force trajectory also closely follows that of the MD simulations, and the failure is correctly brittle. In all settings the total force increases monotonically to its peak and then quickly drops to zero. When $L=50\,lu$, the peak transmitted force in MD is $572\,fu$ at $t=94\,tu$, and the MNO captures the same peak time with a predicted peak force of $595\,fu$. When the domain size and the applied displacement both increase, the peak transmitted force in MD increases by $62.24\%$, while that in the MNO increases by $63.53\%$, corresponding to a $4.83\%$ error in the predicted peak force ($973\,fu$ in MNO vs $928\,fu$ in MD) and an $8.12\%$ error in the peak time ($t=69\,tu$ in MNO vs $t=66\,tu$ in MD). When increasing the internal length ${\tilde{\delta}}$ from $15.05\,lu$ to $30.10\,lu$, the total force trajectory remains consistent, with a peak of $879\,fu$ at $t=67\,tu$. This result demonstrates that the learned material law transfers across horizons with only a mild ${\tilde{\delta}}$-sensitivity of the failure point.

\begin{figure}[t!]
  \centering
  \includegraphics[width=0.49\linewidth]{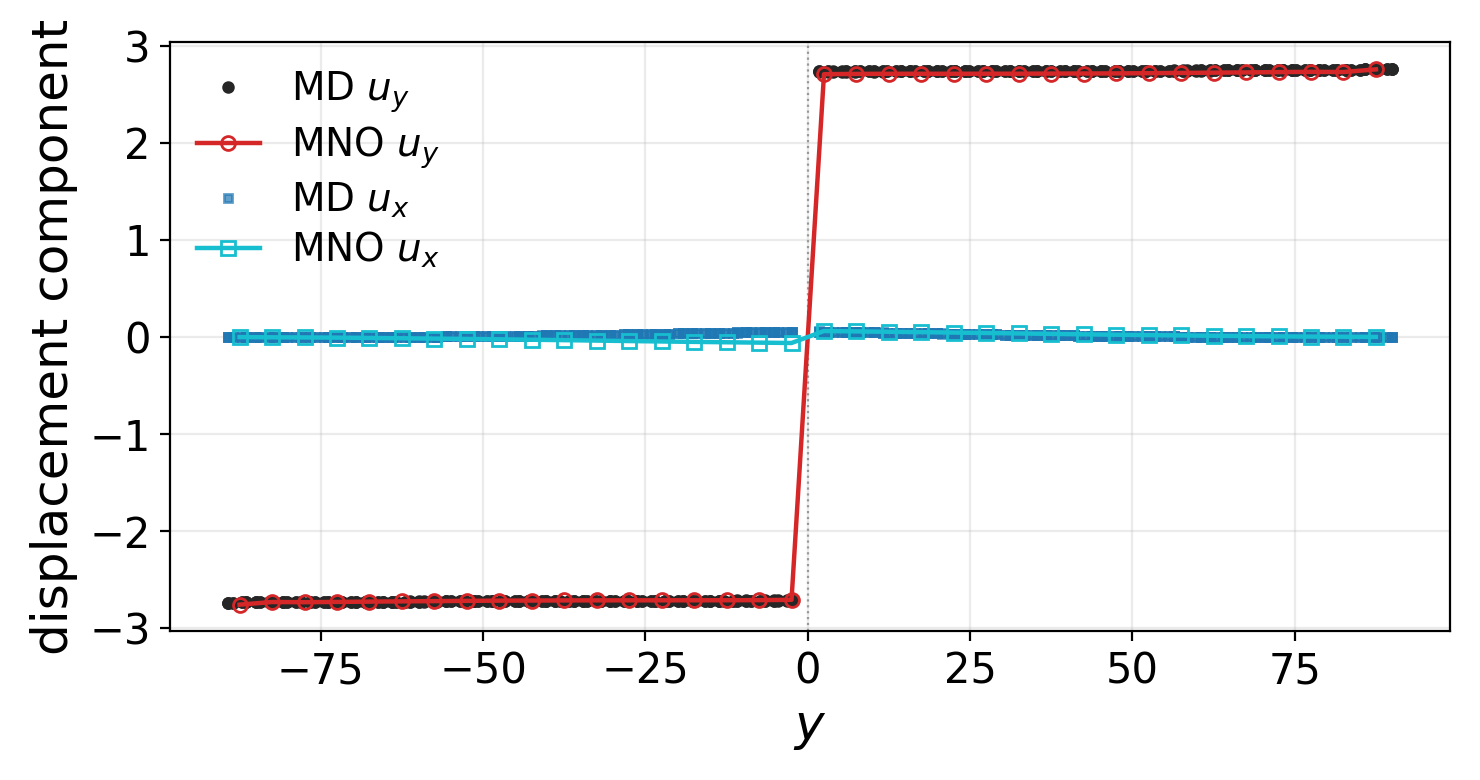}
  \includegraphics[width=0.49\linewidth]{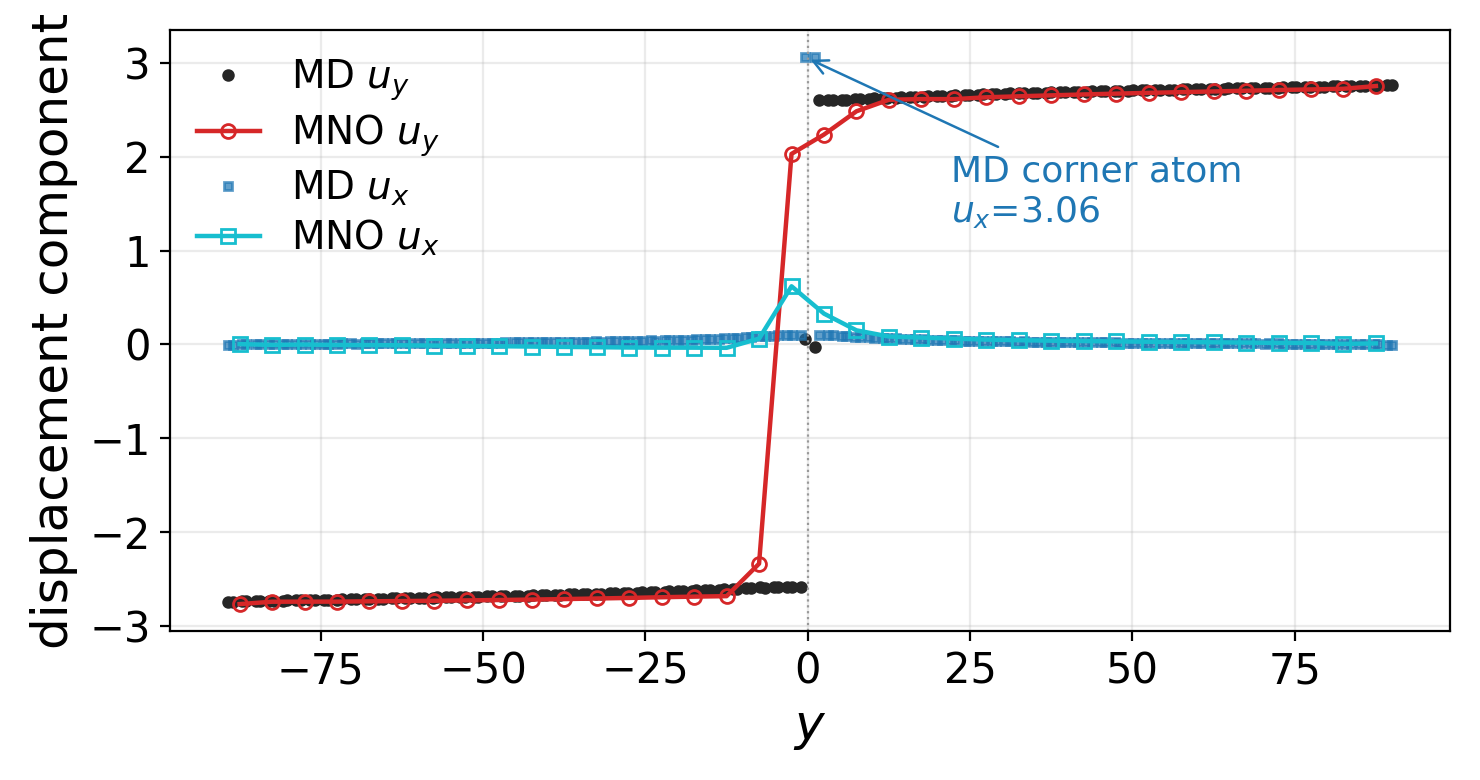}
  \caption{Comparison of MD and MNO final displacement profiles along the center line ($x=0$, left plot) and the free edge ($x=-L$, right plot) in the $100\times 200$ pre-notched plate case. One can see that in the middle of the plate the MNO and MD predictions are very consistent, whereas at $x=-L$ the discrepancy concentrates at the crack corner, where the solution is unstable as fracture occurs.}
  \label{fig:fdivh_final_profiles}
\end{figure}

\textbf{Comparison of the displacement fields.} In addition to the total force, we compare the full displacement fields against the measured MD particle data at two  time instants: the force peak where fracture starts, and the final time where the plate is fully separated into two halves.  The MD displacements are interpolated onto the MNO grid nodes $\chi$. To ensure that the displacement discontinuity across the crack is not smeared out, we employ the weighted essentially non-oscillatory (WENO)-based coarse-graining method developed in \cite{you2023towards}. This is an adaptive procedure that automatically chooses the locally smoothest stencil to reconstruct the coarse-grained displacement field from the MD displacements as a piecewise-smooth solution containing discontinuities. We then report the relative $L^2(\Omega)$ error for the displacement fields as
\begin{equation}\label{eq:disp_error}
  e(t) \;=\;
  \frac{{\|\ub^{MNO}(t)-\ub^{MD}(t)\|_{L^2(\Omega)}}}
       {\|\ub^{MD}(t)\|_{L^2(\Omega)}}.
\end{equation}
The resulting errors are collected in Table~\ref{tab:disp_errors}. Figures~\ref{fig:particles-nofracture}--\ref{fig:particles-L100} show the corresponding fields and pointwise errors for non-fracture plate, fracture plate with $L=50$, and fracture plate with $L=100$, respectively. For the non-fracture case, the final displacement fields match very well: the relative $L^2$ error is only $1.32\%$. For the pre-notched cases, at the fracture peak the displacement fields agree to $2.97\%$ ($L=50$) and $3.84\%$--$4.00\%$ ($L=100$), also showing consistent predictions before fracture propagates. At the final state, the displacement errors increase to $9.43\%$--$15.79\%$. As shown in Figures \ref{fig:particles-L50}--\ref{fig:particles-L100}, the displacement errors are dominated by the discrepancy at the crack interface and a few tethered nodes at the two corners, where the crack meets the lateral free surfaces. Excluding these regions reduces the final-state relative displacement errors to $1.92\%$, $2.37\%$, and $3.32\%$ for Settings 1--3, respectively. In Figure \ref{fig:fdivh_final_profiles} we demonstrate the displacement profiles at the center line ($x=0$) and at the left side ($x=-L$) of the plate. One can see that at $x=0$ the MNO reproduces both displacement components essentially exactly. 
The larger errors are concentrated at the free edges $x=\pm L$, where the cracks reach the traction-free surfaces. In these regions, the MNO profile becomes slightly asymmetrical.
This asymmetry reflects the inherently unstable nature of dynamic fracture, which causes a crack trajectory to oscillate and sometimes branch (see Section~\ref{sec:downstream} for a more dramatic example of this). 
Looking into the details of these errors,
the MNO places the opening jump one node spacing below the crack plane, 
under-opens the lower crack face near the mouth, and develops a lateral displacement of $u_x=0.62\,lu$ at the corner node.
This is roughly six times the MD value of $u_x=0.10\,lu$. On the other hand, the MD solution itself contains two atoms that detach from the lattice and slide along the free surface to $u_x=3.06\,lu$. These profiles indicate that the residual global displacement error stems chiefly from the solution instability when fracture occurs, and that this error is amplified as the crack propagates toward the two ends of the plate.

\subsection{Downstream Crack-Branching Simulation at $\tilde{\delta}=10\delta$}\label{sec:downstream}

  \begin{figure}[t!]
   \centering
   \includegraphics[width=0.7\textwidth]{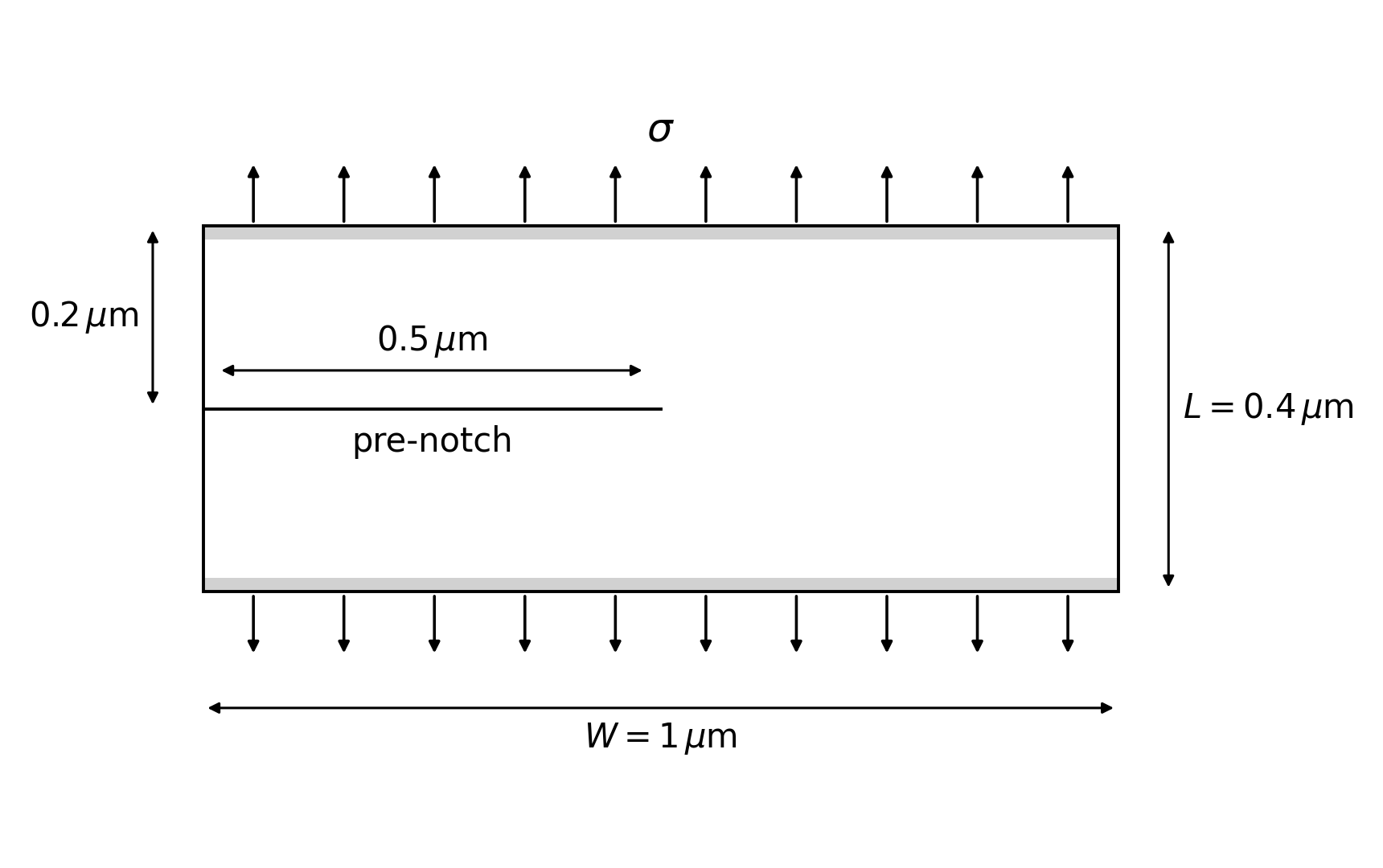}
   \caption{Problem setup for pre-notched crack branching experiment under tensile loading, following \cite{ha2010studies,yu2021asymptotically}.}
   \label{fig:glasssetup}
 \end{figure}

 \begin{table}[h!]
\center
\begin{tabular}{l|cc}
\hline
Parameter & SI units & MD units\\
\hline
Grid spacing $\Delta x$ & $5\times 10^{-9}\,m$ & $50\,lu$\\
Time step $\Delta t$ & $1\times 10^{-13} \sim 1\times 10^{-12}\,s$ & $0.1\sim 1\,tu$\\
Density $\rho$ & $8000\ kg/m^3$ & $8\, mu/lu^{3}$\\
Domain length $L$ & $4.05\times 10^{-7}\,m$ & $4050\, lu$\\
Domain width $W$ & $1\times 10^{-6}\,m$ & $10000\, lu$\\
Plate thickness $H$ & $9.35\times 10^{-11}\,m$ & $0.935\,lu$\\
Total simulation time $t_{\rm f}$ & $5\times 10^{-9} \sim 5\times 10^{-8}\,s$ & $5\times 10^3 \sim 5\times 10^4\,tu$\\
Horizon $\tilde{\delta}$ & $1.505\times 10^{-8}\,m$ & $150.5\, lu$\\
Traction Load $\sigma$ & $1.07\sim 21.4\,MPa$ & $0.107\sim 2.14\,su$\\
\hline
\end{tabular}
\caption{Physical parameters of the downstream crack-branching simulation at $\tilde{\delta} = 10\delta$, in both International System of Units (SI) and MD system units.}
\label{tab:branching_params}
\end{table}

Finally, we employ the learned model in a downstream dynamic-fracture application at a length scale far beyond the reach of the underlying MD. We simulate crack propagation and branching in a pre-notched thin plate under sudden tensile loading. As shown in Figure \ref{fig:glasssetup}, a thin plate of width $W = 1\,\mu\mathrm{m}$, height $L = 0.405\,\mu\mathrm{m}$, and thickness $H=9.35 \times 10^{-5} \,\mu\mathrm{m}$, contains an edge pre-notch of length $0.5\,\mu\mathrm{m}$ at mid-height. The density is $\rho = 8\ mu/lu^3 = 8000\ kg/m^3$.  Starting at time zero, a constant tensile traction $\sigma$ is applied on the top and bottom faces of the plate. This geometry has been investigated in several numerical studies at different length scales \cite{ha2010studies,gu2017voronoi,du2017peridynamic,zhou2016numerical,diehl2019review}. In these studies, complex crack branching behavior is observed. As $\sigma$ increases, the crack develops branches and sub-branches. Although the geometrical dimensions considered here are at the micron scale, this is much larger than the length scale in MD. This problem therefore assesses the ability of the MNO to predict complex branching behaviors in downstream applications at a much larger scale.

\begin{figure}[p]
  \centering
  \includegraphics[width=0.95\linewidth]{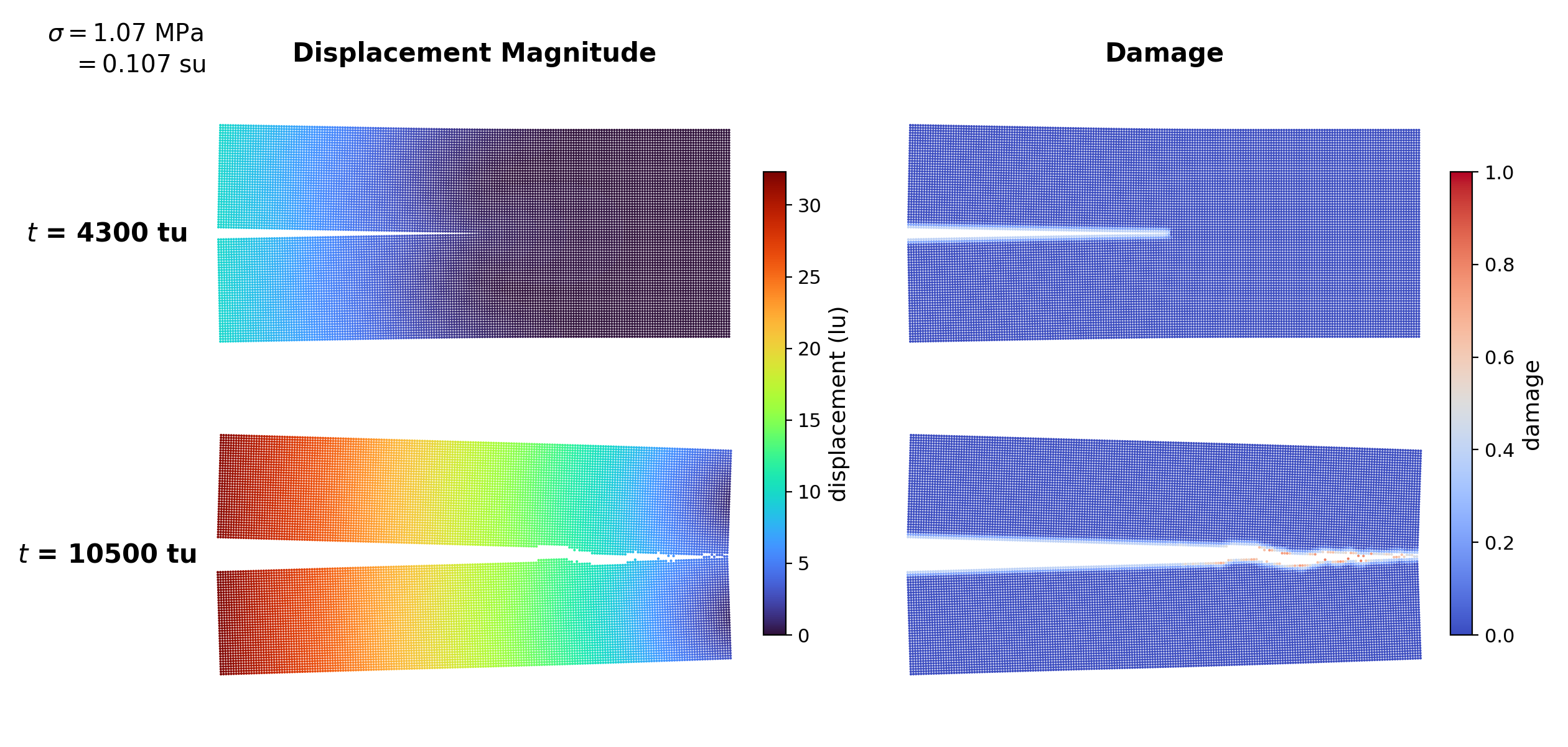}
  \caption{Downstream crack-branching simulation at
  $\tilde\delta = 10\delta$ under the total load $T = 10^3\,fu$
  ($\sigma = 1.07\,MPa = 0.107\,su$): displacement magnitude
  $\sqrt{u_x^2+u_y^2}$ (left column) and damage (right column) on the
  deformed configuration at crack nucleation and full fracture. A single crack
  propagates straight across the plate.}
  \label{fig:branching_T1e3}
\end{figure}

\begin{figure}[p]
  \centering
  \includegraphics[width=0.95\linewidth]{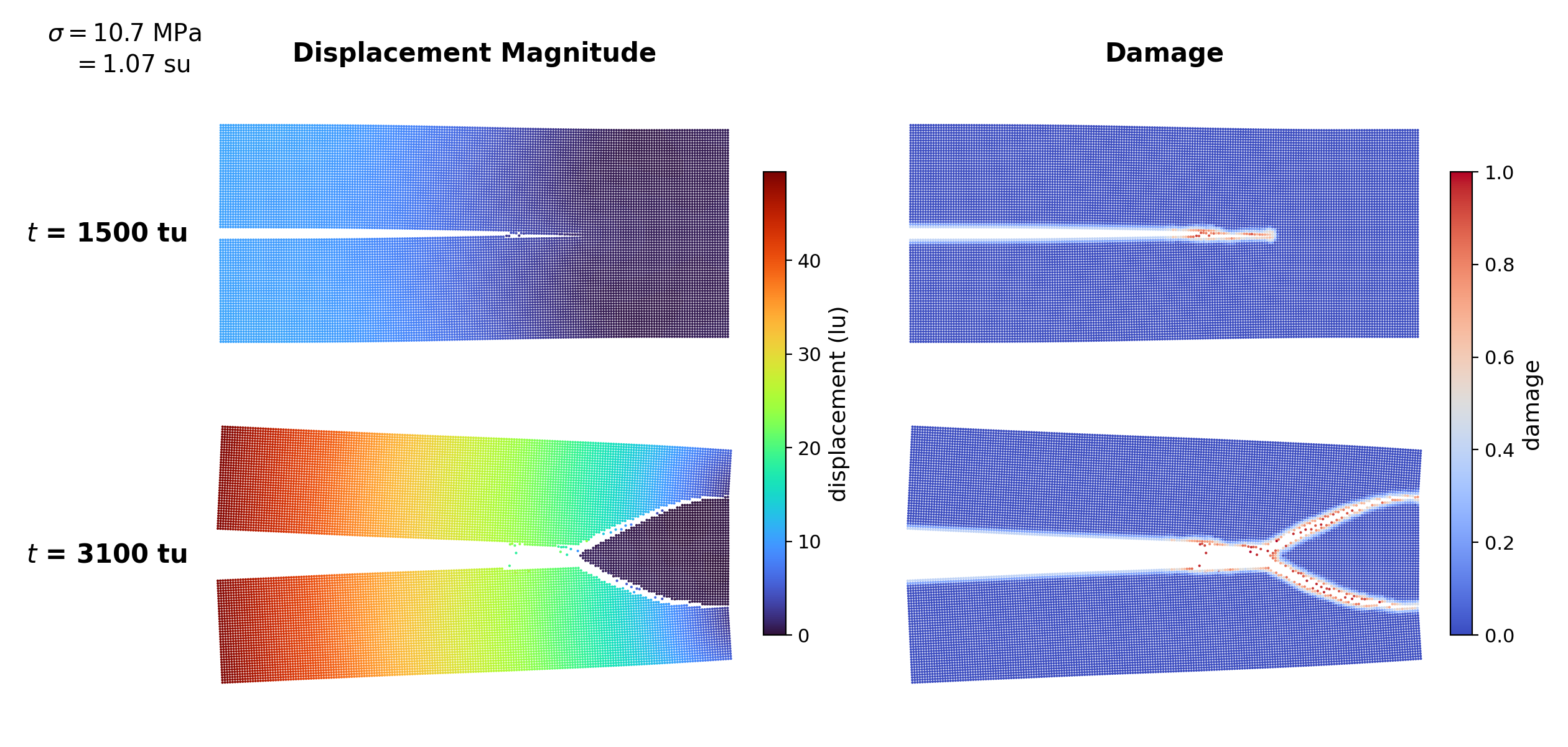}
  \caption{Downstream crack-branching simulation at
  $\tilde\delta = 10\delta$ under the total load $T = 10^4\,fu$
  ($\sigma = 10.7\,MPa = 1.07\,su$): displacement magnitude
  $\sqrt{u_x^2+u_y^2}$ (left column) and damage (right column) at crack bifurcation and full fracture. The
  crack runs straight through the first half of the plate and then
  bifurcates into an almost symmetric Y-shaped branch pair.}
  \label{fig:branching_T1e4}
\end{figure}

\begin{figure}[p]
  \centering
  \includegraphics[width=0.95\linewidth]{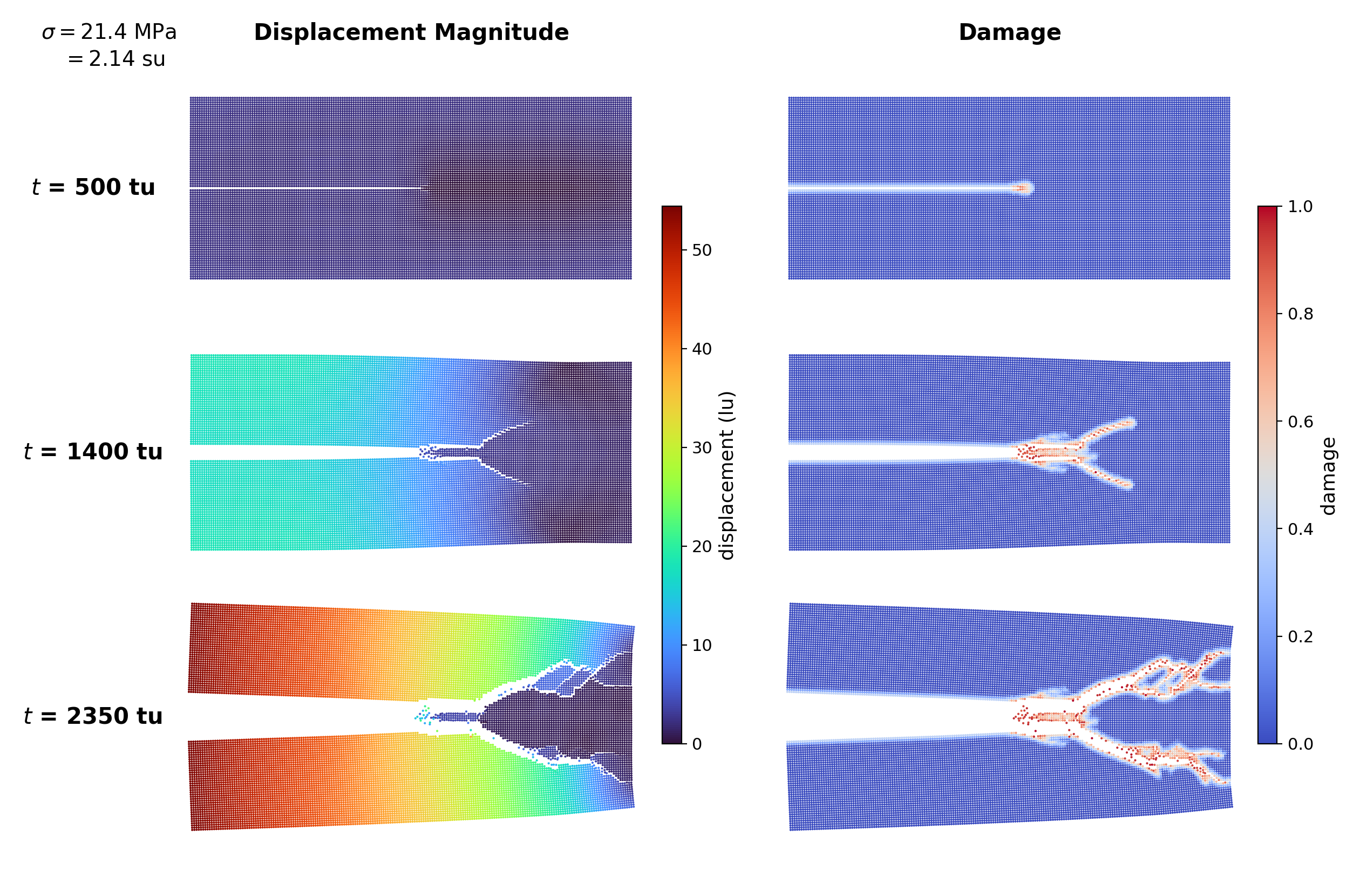}
  \caption{Downstream crack-branching simulation using MNO at
  $\tilde\delta = 10\delta$ under the total load $T = 2\times 10^4\,fu$
  ($\sigma = 21.4\,MPa = 2.14\,su$): displacement magnitude
  $\sqrt{u_x^2+u_y^2}$ (left column) and damage (right column) at crack nucleation, first branching, and final fracture.  The primary branches themselves branch, producing a non-symmetric cascading
  crack network that spans the right half of the plate.}
  \label{fig:branching_T2e4}
\end{figure}

The area of the plate is
approximately $4000$ times the area of the original training specimens in Section \ref{sec:MD}.  To make this scale tractable, the learned material model is upscaled to the horizon size $\tilde\delta = 10\delta = 150.5\,lu$ following \eqref{eq:nonlocElasticity_comp_tilde}, and discretized with
$\Delta x = 50\,lu$, so that the domain is discretized into $201\times 82$ nodes with $\tilde\delta/\Delta x = 3.01$. The traction loading boundary condition is applied as a body force density $\bb = (0,\, \pm\sigma/\tilde\delta)$ over the $\tilde\delta$-thick strips along the top and bottom faces (shaded in Figure \ref{fig:glasssetup}). This results in a total load $T=\sigma W H$. All remaining boundaries, including the newly created crack faces, are free surfaces.  The dynamic MNO equation is integrated with an implicit Newmark scheme, with Newton's method as the nonlinear solver at each step.

Figures~\ref{fig:branching_T1e3}--\ref{fig:branching_T2e4} show the evolution of the displacement and damage fields for traction loads $\sigma = 1.07$, $10.7$, and $21.4\, MPa$, plotted on the deformed configuration $\xb + \ub$.  The model reproduces the classical progression of dynamic fracture morphologies with load amplitude: 
\begin{itemize}
\item At the lowest load, $\sigma = 1.07\,MPa$, the plate at first stores elastic strain before the crack starts growing ($t = 4300\,tu$). The crack then initiates and propagates straight across the plate. The end result is that it symmetrically separates the two halves ($t = 10500\,tu$). 
\item At the intermediate load, $\sigma = 10.7\,MPa$, the crack initiates and accelerates over the first half of the plate ($t = 1500\,tu$). It then bifurcates into an almost symmetrical Y-shaped branch pair ($t = 3100\,tu$), showing the hallmark instability of overdriven brittle cracks. 
\item At the highest load, $\sigma = 21.4\,MPa$, a damage process zone develops at the notch tip almost immediately ($t = 500\,tu$), and the primary bifurcation is almost symmetric ($t = 1400\,tu$). These initial branches themselves branch repeatedly, producing a non-symmetric cascading crack network that sweeps the right half of the plate until it reaches the far edge ($t = 2350\,tu$).
\end{itemize}
As would be expected, the time to full fracture decreases when the applied load is increased, from roughly $10^4\,tu$ at $\sigma = 1.07\,MPa$ to roughly $2\times 10^3\,tu$ at $\sigma = 21.4\,MPa$, while the crack surface created grows from a single flat cut to a hierarchical branch network. This progression from flaw-controlled growth of a single crack to strength-controlled nucleation with profuse branching emerges entirely from the learned constitutive law and the rescaled damage criterion, and it is obtained at a scale that is not feasible for direct MD simulation.

\section{Conclusion}\label{sec:conclusion}

In this work we proposed the mesoscale neural operator (MNO) to
bridge the gap between molecular dynamics (MD) simulation, which
captures the microscale physics of fracture but does not scale up to
application-relevant domains, and a machine learning surrogate that
scales up readily. Importantly, our method guarantees the physical consistency of the
predicted response.  The MNO is a data-driven constitutive operator 
developed based on the bond-based peridynamic formulation and learned directly from MD displacement and
body-load data. The built-in characteristic length of the
nonlocal formulation, the horizon $\delta$, serves as the vehicle for
traveling across scales.  Physical consistency across scales is
obtained through two structural ingredients.  First, a
normalization of the learned kernel with respect to the horizon
$\delta$ renders the strain energy density invariant under horizon
rescaling.  Second, a rescaled critical bond stretch
$\lambda_0(\tilde\delta)$ keeps the energy release rate
consistent when the learned model is transferred from the training
horizon $\delta$ to a larger simulation horizon $\tilde\delta$.
Together, these two properties guarantee that the energetic quantities
governing crack growth are preserved across scales for objects under affine deformation.

We first verified the learning framework on synthetic data sets
generated at two characteristic lengths, where the learning workflow
discovered the ground-truth kernel and nonlinear constitutive
relation.  We then applied
the framework to learn from coarse-grained quasi-static MD
simulations of a two-dimensional plate at the angstrom
scale.  The learned model predicted held-out displacement fields
within $10\%$ relative error and, more importantly, reproduced the two
energy-based quantities computed directly from the MD potential, the
strain energy density and the energy release rate, within $3\%$ without supervision.  
The model was further validated through a downstream application of
displacement-controlled tension tests, in which the domain and loading scenarios lie outside the training
distribution. We compared the predicted peak transmitted forces and displacement fields
with the results from MD simulations and found that the MNO predictions matched MD within
$6\%$ across domain sizes. Finally, we upscaled the learned model to
$\tilde\delta = 10\delta$ and simulated dynamic fracture of a
pre-notched plate with $4000$ times the area of the training
specimens.  The simulations reproduced the classical progression of
brittle-fracture morphologies with load amplitude, from the straight
growth of a single crack, to symmetric Y-shaped bifurcation, to
cascading branch networks.

Several limitations suggest directions for future work.  The present
model is developed based on bond-based peridynamics. Extending the normalization and critical-stretch
rescaling to state-based formulations such as the peridynamic neural
operator \cite{jafarzadeh2023peridynamic} would further relax
the restrictions on the Poisson ratio and material symmetry.  The damage rule remains a critical-stretch criterion
inherited from classical peridynamics, and the possibility of a continuously varying damage variable for
bonds has not been addressed here. Finally, both the thermal noise in MD simulations and the instability of 
cracks unavoidably introduce errors and uncertainties.
Transferring learned models across temperatures and loading rates, and
quantifying the uncertainty of the learned constitutive form, would
further strengthen the MD-to-continuum pipeline proposed here.

\section*{CRediT authorship contribution statement}

\textbf{Yue Yu:} Conceptualization, Methodology, Software, Analysis, Writing -- original draft.

\textbf{Jihong Wang:} Methodology, Software, Writing -- original draft.

\textbf{Stewart Silling:} Data, Analysis, Editing.

\textbf{Ziwei Yang:} Methodology.

\section*{Declaration of competing interest}
The authors declare that they have no known competing financial interests or
personal relationships that could have appeared to influence the work reported
in this paper.

\section*{Data availability}
Datasets and source codes are available on \url{https://github.com/fishmoon1234/PeridynamicNeuralOperator}.

\section*{Acknowledgements}
Authors acknowledge support by the Defense Advanced Research Projects Agency (DARPA) under award HR00112590032. This research was, in part, funded by the U.S. Government by an agreement with Cornell University.  Portions of this research were conducted on Lehigh University's Research Computing infrastructure partially supported by NSF Award 2019035.


\bibliographystyle{elsarticle-num}
\bibliography{yyu}

\end{document}